%% file: main.tex
\documentclass[10pt,twocolumn,letterpaper]{article}
\usepackage[pagenumbers]{iccv} 

\definecolor{iccvblue}{rgb}{0.21,0.49,0.74}
\usepackage[pagebackref,breaklinks,colorlinks,allcolors=iccvblue]{hyperref}

\usepackage{threeparttable}
\usepackage{multirow}
\usepackage{tabularx}
\usepackage{ragged2e}
\usepackage{booktabs}
\usepackage{makecell}
\usepackage{xspace}
\usepackage{diagbox}
\usepackage{xcolor}
\usepackage{soul}
\usepackage{amssymb}
\usepackage{comment}
\usepackage{balance}	
\usepackage{color, colortbl}
\usepackage{tikz}
\usetikzlibrary{fit,calc}

\colorlet{pink}{red!40}

\usepackage{algorithm}
\usepackage{algpseudocode}
\usepackage{ragged2e}
\usepackage{hyperref}
\usepackage{xspace}
\newcolumntype{C}[1]{>{\centering\arraybackslash}p{#1}}
\newcolumntype{P}[1]{>{\centering\arraybackslash}p{#1}}
\newcolumntype{M}[1]{>{\centering\arraybackslash}m{#1}}
\newcolumntype{L}[1]{>{\raggedright\arraybackslash}p{#1}}
\newcolumntype{R}[1]{>{\raggedleft\arraybackslash}p{#1}}
\newcolumntype{J}[1]{>{\justifying\arraybackslash}p{#1}}

\usepackage{tikz}
\usepackage{comment}
\newcommand{\papersource}[1]{\tiny{\color{blue}{#1}}}
\newcommand{\mypara}[1]{\noindent\textbf{#1}}
\newcommand{\myMethod}{\mbox{CoDa-4DGS}\xspace} 
\makeatletter
\def\blfootnote{\xdef\@thefnmark{}\@footnotetext}

\title{CoDa-4DGS: Dynamic Gaussian Splatting with Context and Deformation Awareness for Autonomous Driving}

\author{
    Rui Song\textsuperscript{\rm 1,\rm 2 *},
    Chenwei Liang\textsuperscript{\rm 1 *},
    Yan Xia\textsuperscript{\rm 2},
    Walter Zimmer\textsuperscript{\rm 2},
    Hu Cao\textsuperscript{\rm 2}\\
    \vspace{1ex}
    Holger Caesar\textsuperscript{\rm 3},
    Andreas Festag\textsuperscript{\rm 1,\rm 4},
    Alois Knoll\textsuperscript{\rm 2}\\
    $^{1}$Fraunhofer IVI \quad $^{2}$TU Munich \quad $^{3}$TU Delft \quad $^{4}$TH Ingolstadt
}

\begin{document}
\twocolumn[{%
\renewcommand\twocolumn[1][]{#1}%
\maketitle

\begin{center}
    \centering
    \captionsetup{type=figure}
    \includegraphics[trim={1.6cm 15cm 2cm 1cm},clip, width=1\textwidth]{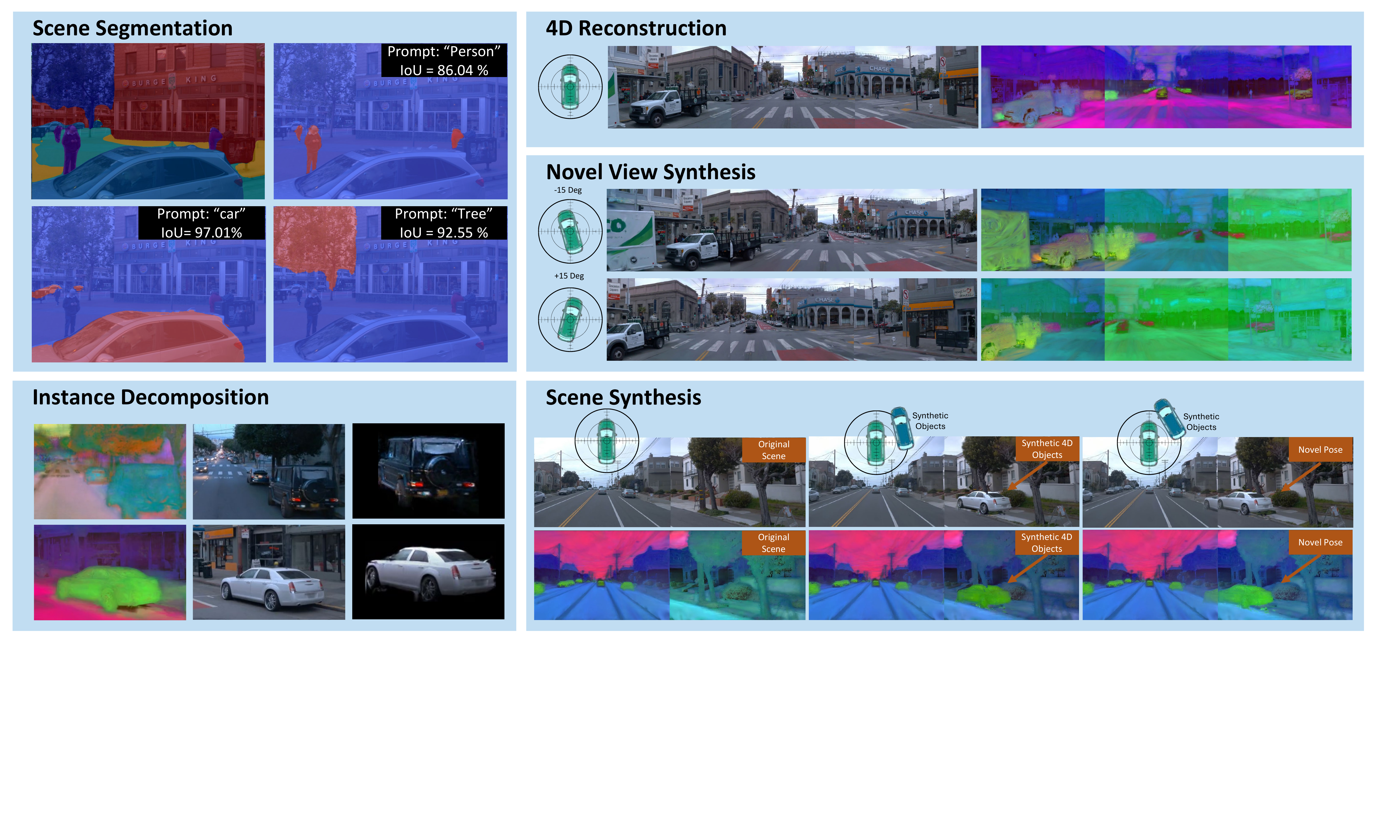}
    \captionof{figure}{Our CoDa-4DGS incorporates both context awareness and deformation awareness to effectively compensate for deformable Gaussians in 4D. This results in more accurate dynamic scene rendering and enables a range of downstream applications, such as scene segmentation, instance segmentation, 4D reconstruction, novel view synthesis, and scene synthesis. Note that we use Principle Component Analysis (PCA) to visualize the splatted context awareness for each camera next to their RGB rendered results. More application demonstrations are available in the supplementary material.}
    \label{fig:teaser}
\end{center}%
}]
\blfootnote{* Equal contribution}

\input{sec/0_abstract}    
\input{sec/1_intro}
\input{sec/2_related_work}

\input{sec/3_method}
\input{sec/4_experiments}
\input{sec/5_conclusion}

\onecolumn
\newpage
\twocolumn

{
    \small
    \bibliographystyle{ieeenat_fullname}
    \bibliography{main}
}

\onecolumn
\newpage
\twocolumn

\maketitlesupplementary

In this supplementary material, we provide additional implementation details in Sec.~\ref{sec:imp_details}. In Sec.\ref{sec:plugplay}, we showcase the plug-and-play functionality of our approach, demonstrating how CoDa-4DGS enhances the performance of both vanilla 4DGS~\cite{wu20244d} and $S^3$Gaussian~\cite{huang2024textit}. Sec.~\ref{sec:scene_editing} presents visualization results for 4D dynamic scene editing, highlighting the distinctions between our method and prior work that predominantly focuses on 3D. In Sec.~\ref{sec:nvs}, we provide visual results for novel view synthesis, addressing scenarios with large ego-view angle shifts, thereby extending beyond previous evaluations that primarily consider small frame transitions in test sets. Additionally, Sec.~\ref{sec:discussion} offers an in-depth conceptual comparison with recent works, while Sec.~\ref{sec:further_vis} includes extended 4D visualizations to demonstrate the robustness and versatility of our approach.

\appendix
\input{sec/x1_imp_details}    
\input{sec/x2_plug_and_play}
\input{sec/x3_scenario_editor}
\input{sec/x4_novel_view_synthesis}
\input{sec/x5_discussion}
\input{sec/x6_further_4Dvis}

\end{document}

%% file: sec/0_abstract.tex
\begin{abstract}

Dynamic scene rendering opens new avenues in autonomous driving by enabling closed-loop simulations with photorealistic data, which is crucial for validating end-to-end algorithms. However, the complex and highly dynamic nature of traffic environments presents significant challenges in accurately rendering these scenes.
In this paper, we introduce a novel 4D Gaussian Splatting (4DGS) approach, which incorporates context and temporal deformation awareness to improve dynamic scene rendering.
Specifically, we employ a 2D semantic segmentation foundation model to self-supervise the 4D semantic features of Gaussians, ensuring meaningful contextual embedding. 
Simultaneously, we track the temporal deformation of each Gaussian across adjacent frames. 
By aggregating and encoding both semantic and temporal deformation features, each Gaussian is equipped with cues for potential deformation compensation within 3D space, facilitating a more precise representation of dynamic scenes.
Experimental results show that our method improves 4DGS's ability to capture fine details in dynamic scene rendering for autonomous driving and outperforms other self-supervised methods in 4D reconstruction and novel view synthesis. 
Furthermore, CoDa-4DGS deforms semantic features with each Gaussian, enabling broader applications.
\end{abstract}

%% file: sec/1_intro.tex
\section{Introduction}
\label{sec:intro}

Validations in autonomous driving have been largely based on virtual scenarios generated by rendering engines or simulators, such as CARLA~\cite{dosovitskiy2017carla}. While real-world dataset collection is effective for verifying perception algorithms, it presents challenges for the emerging end-to-end autonomous driving~\cite{hwang2024emma}, where decision-making and planning are derived directly from visual inputs. Offline validation using real-world data is difficult in these cases because vehicle trajectories, sensor poses, and associated scenes are fixed to the moments of data collection, limiting flexibility. This constraint prevents thorough validation of the new trajectories generated by autonomous driving functions, making comprehensive assessment challenging.

Recently, with the rise of neural rendering and novel view synthesis technologies, the validation of autonomous driving based on photorealistic data has gained a new foundation. By collecting a small set of images from different perspectives, 3D scene information can be represented within a model, enabling new view generation and editing of the original scene from varied angles. This advancement makes closed-loop simulation~\cite{yang2023unisim, ljungbergh2024neuroncap} for autonomous driving with real data feasible, specifically allowing: (\emph{i}) precise new view rendering of a scene based on different trajectories generated by various algorithms; and (\emph{ii}) adjustments to real scenes as needed for validation, simulating more diverse scenarios, including more critical ones. This marks a revolutionary step forward in the development and validation of autonomous driving.

Mainstream neural scene rendering techniques, such as NeRF~\cite{rudnev2022nerf} and Gaussian splatting~\cite{kerbl3Dgaussians}, were initially designed for rendering static 3D scenes. For dynamic scenes, these methods typically rely on merging multiple 3D scenes to capture changes over time. To overcome limitations in model size and training time with this approach, an increasing number of 4D scene rendering techniques have been introduced. These techniques allow dynamic scene rendering within a single model by learning temporal changes directly, enabling more efficient rendering of time-varying scenes.

However, unlike other applications, dynamic scene rendering for autonomous driving poses unique challenges. First, autonomous driving environments are typically expansive, often covering hundreds of meters, and are characterized by intricate, detailed backgrounds. They include a diverse mix of large structures like buildings and vehicles, as well as smaller yet critical elements such as traffic signs and road markers, all of which contribute to their complexity. Secondly, static and dynamic objects coexist. Moving traffic participants, such as cars, bicycles, and pedestrians, introduce significant temporal changes. This movement amplifies the impact of dynamic objects on the quality of scene rendering. Third, data collected for autonomous driving are relatively limited. Image acquisition is influenced by factors such as vehicle trajectories, speed, and capture frame rate, often leading to less varied data compared to scenarios like indoor environments, which are typically used for dynamic rendering. This restricted diversity in supervised training data for scene models further complicates the challenge of building robust scene representations for autonomous driving.

To address the challenges of dynamic scene rendering in autonomous driving, we propose a \textbf{Co}ntext- and temporal \textbf{D}eformation-\textbf{a}ware \textbf{4D} \textbf{G}aussian \textbf{S}platting (\myMethod) method. Previous methods achieved time-varying scene rendering by learning the temporal deformation of Gaussians. However, this approach can be inadequate for newly introduced dynamic objects in the scene, as these Gaussians may lack access to relevant information from prior frames, hindering effective learning of Gaussian deformation. To solve this common challenge in autonomous driving scenarios, as shown in Fig.~\ref{fig:teaser}, we incorporate two types of awareness in training: (\emph{i}) Context awareness: inspired by Feature-3DGS~\cite{zhou2024feature}, we leverage a 2D foundational model, such as LSeg~\cite{li2022language} or Meta-SAM~\cite{kirillov2023segany}, to supervise each Gaussian in learning 4D semantic information, and (\emph{ii}) temporal deformation awareness, derived from learned deformation information over time. Leveraging these additional awareness dimensions, we predict a deformation compensation to Gaussians at each time step. Our experiments show that this compensated deformation effectively refines 3D reconstruction details and enhances overall rendering quality.

\begin{figure*}[t!]
   \centering
   \includegraphics[trim={6cm 5.5cm 8cm 2cm},clip, width=1\textwidth]{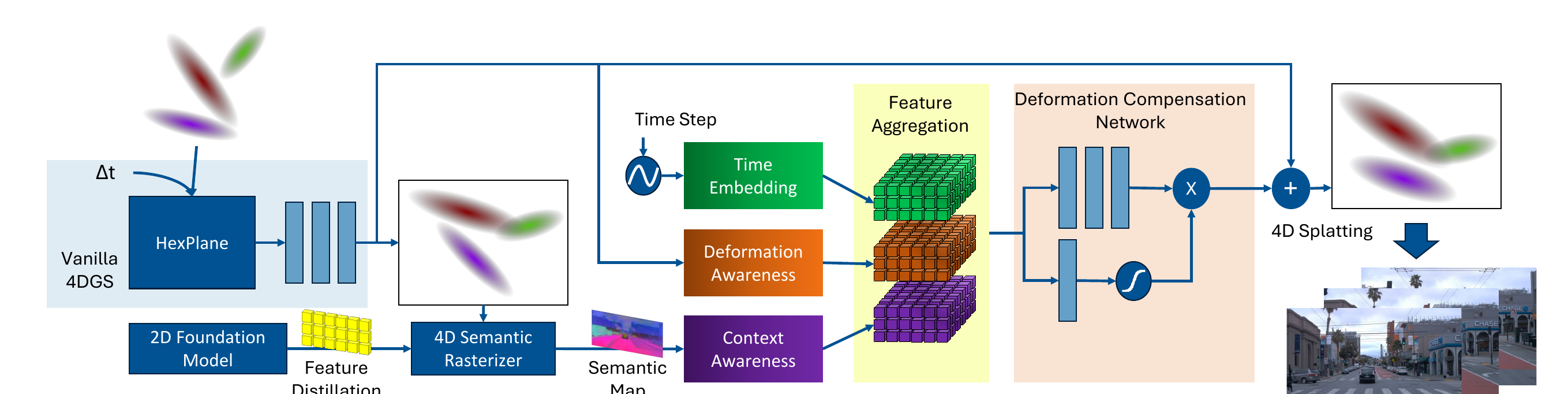}
   \caption{System overview of \myMethod. Vanilla 4DGS~\cite{wu20244d} encodes and decodes temporal deformation using HexPlane~\cite{cao2023hexplane} encoding. Building on this, our \myMethod embeds temporal information and aggregates it with context and temporal deformation awareness through a Deformation Compensation Network (DCN). This network encodes the deformation adjustments needed to compensate for the original temporal deformation, ultimately producing an enhanced set of 4D Gaussians.}
   \label{fig:system}
\end{figure*}

\noindent\textbf{Contributions}. To summarize, our main contributions are threefold:
\begin{itemize}
    \item We introduce context and temporal deformation awareness for Gaussians. Context awareness is generated by distilling 4D semantic features based on a 2D foundational model, providing contextual information for 4D Gaussians. Using the temporal deformation of 4D Gaussians, we further adjust their deformation within 3D space at each time step.
    \item We propose the \myMethod, which aggregates context and temporal deformation awareness information to train a network module for compensating Gaussian deformation. We integrated a Deformation Compensation Network (DCN) that effectively compensates for the deformation of Gaussians generated by the spatiotemporal structure encoder, enhancing the model's performance in dynamic scenarios. 
    \item We evaluate and validate the effectiveness of the \myMethod on various autonomous driving scenes. Our method can be used independently or as a plug-and-play enhancement for existing 4DGS models, improving dynamic scene rendering quality in complex scenarios. In addition, we demonstrate instance decomposition of the \myMethod method, highlighting its potential. 
\end{itemize}

%% file: sec/2_related_work.tex
\section{Related work}
\label{sec:related_work}

In this section, we review dynamic scene rendering in~\ref{sec:dyn}. In Sec.~\ref{sec:nerf4ad} and Sec.~\ref{{sec:gs4ad}}, we explore NeRF- and GS-based approaches for rendering in autonomous driving scenes.

\subsection{Dynamic scene rendering}
\label{sec:dyn}

Dynamic scene representation enables scene rendering with a temporal dimension. In addition to NeRF-based dynamic scene rendering methods~\cite{li2021neural, park2021nerfies,park2021hypernerf,pumarola2021d,tretschk2021non, xian2021space, lombardi2021mixture, li2022neural, liu2022devrf, lin2022efficient, fang2022fast, Chen2022ECCV, song2023nerfplayer, peng2023representing, guo2023forward, li2023dynibar, lin2023im4d, wang2023mixed,wang2023neural, attal2023hyperreel, xie2023snerf, sun2024lidarf}, Gaussian-based approach has also attracted considerable attention. Previous work, such as~\cite{luiten2024dynamic, yang2024deformable, Li_STG_2024_CVPR, huang2024sc, zhu2024motiongs, yang2024gs4d, wu20244d,10816486,NEURIPS2024_93397b48} has achieved high-quality dynamic scene rendering. Dynamic-3DGS~\cite{luiten2024dynamic} reconstructs and renders dynamic scenes by tracking 3D Gaussians over time. Deformable-3DGS~\cite{yang2024deformable} uses an MLP deformation network to model Gaussian motion across temporal sequences. Spacetime-GS~\cite{Li_STG_2024_CVPR} further builds on 3D Gaussian tracking by capturing each Gaussian’s temporal opacity, polynomial motion, rotation, and other time-dependent features. The approach in~\cite{yang2024gs4d} achieves a cohesive integration of spatial and temporal dimensions via unbiased 4D Gaussian primitives. 

4DGS~\cite{wu20244d} enhances the modeling efficiency for dynamic scenes by using HexPlane~\cite{cao2023hexplane, fridovich2023k} encoding along with an MLP to predict Gaussian deformations at novel timestamps. Our work extends these approaches by introducing deformation awareness through a sigmoid-based mask network, enabling a more precise refinement of the predicted deformations. Recently, DN-4DGS~\cite{lu2024dn} incorporates a noise suppression strategy and a decoupled temporal-spatial aggregation module to assimilate information from neighboring points and frames. Unlike DN-4DGS, which requires two separate MLPs across distinct training stages, our method employs a single model with unified training, simplifying the process.

\subsection{NeRF for autonomous driving}
\label{sec:nerf4ad}

With the growing interest in neural rendering for autonomous driving, particularly for closed-loop simulation and photorealistic dataset synthesis, several methods based on NeRFs have been developed to address the complexities of modeling urban and dynamic scenes~\cite{rematas2022urban, rudnev2022nerf, lu2023urban, liu2023real, lu2023urban, turki2023suds}. Block-NeRF~\cite{tancik2022block} achieves efficient scene representation through block-based segmentation. Mega-NeRF~\cite{turki2022mega} targets large-scale cityscapes for urban air mobility using multi-perspective images, while NSG~\cite{ost2021neural} provides a structured approach to scene complexity by decomposing dynamic scenes into scene graphs. NeuRAD~\cite{tonderski2024neurad} incorporates detailed sensor modeling, including rolling shutter, beam divergence, and ray drop, making it adaptable to cameras and LiDAR. Furthermore, Streetsurf~\cite{guo2023streetsurf} enhances rendering precision by separately processing sub-scenes, and MARS~\cite{wu2023mars} simplifies dynamic scene handling by modeling vehicle objects independently. Moreover, EmerNeRF~\cite{yang2023emernerf} combines static, dynamic, and flow fields to improve the rendering accuracy of dynamic objects. Leveraging NeRF's stable rendering capabilities, NeuroNCAP~\cite{ljungbergh2024neuroncap} introduces a photorealistic closed-loop safety testing framework for autonomous driving.

These methods introduce insightful strategies for handling autonomous driving scenarios, such as separately modeling dynamic objects and incorporating diverse features to enhance scene understanding, inspiring advancements in GS-based approaches.

\subsection{Gaussian splatting for autonomous driving}
\label{{sec:gs4ad}}

Gaussian splatting has proven highly efficient for large-scale scene modeling, making it advantageous for real-time rendering in traffic scenarios over NeRF-based approaches. Hugs~\cite{zhou2024hugs} performs joint optimization of geometry, appearance, semantics, and motion by leveraging a combination of static and dynamic 3D Gaussians. The poses of moving objects are regularized through physical constraints, ensuring realistic motion modeling. For dynamic street scenes, DrivingGaussian~\cite{zhou2024drivinggaussian} leverages a 360° camera and LiDAR to incrementally reconstruct backgrounds while modeling dynamic objects through Gaussian graphs. Very recently, StreetGaussian~\cite{yan2024street} and $S^3$Gaussian~\cite{huang2024textit} extended these insights, improving the rendering of dynamic street scenes with 4DGS. StreetGaussian associates foreground vehicles with background environments, optimizing each object’s point cloud through tracked poses and dynamic spherical harmonics to capture appearance changes. However, it requires labels for dynamic objects, which limits the scalability of its applications. $S^3$Gaussian improves 4D scene neural modeling by decomposing dynamic and static elements in a self-supervised manner.

To summarize, current approaches focus either on precisely identifying moving vehicles or broadly distinguishing between static and dynamic elements. However, in road scenes, certain objects, such as pedestrians, that do not fit neatly into static or dynamic categories present challenges, as the deformation of their Gaussian over time is difficult to predict. Building on previous efforts, we propose to take advantage of 4D semantic features distilled by foundational models, coupled with predicted temporal deformation awareness, to enable secondary neural correction of deformations. This approach would allow for finer adjustments in dynamic scene rendering, capturing detailed variations more accurately.

%% file: sec/3_method.tex
\section{Methodology}
\label{sec:method}

In this section, we introduce CoDa-4DGS and its four key components: context awareness, temporal deformation awareness, time embedding, and a Deformation Compensation Network (DCN), as illustrated in Fig.~\ref{fig:system}.

\subsection{Preliminaries}
\label{sec:pre}
For 3DGS, a given static 3D scene can be represented by $N$ Gaussians (Gaussian blobs). Each Gaussian consists of a 3D centroid $x\in \mathbb{R}^3$ and a 3D covariance matrix $\Sigma\in \mathbb{R}^{3\times3}$, expressed as:

\begin{equation}
G(x) = e^{-\frac{1}{2} x^T \Sigma^{-1} x}.
\end{equation}
where $ \Sigma $ is constructed from a scaling matrix $S$ stored as scaling vector $s \in \mathbb{R}^3$,  and a rotation matrix $R$ stored as a quaternion  $q \in \mathbb{R}^4$, such that

\begin{equation}
\Sigma = R S S^T R^T.
\end{equation}

For novel view synthesis, $\Sigma$ is transformed to $ \Sigma'$ in the 2D image space using a view transform matrix $W$ for the transformation of the world to the camera and a Jacobian matrix $J$ from the affine approximation of the projective transformation:

\begin{equation}
\Sigma' = J W \Sigma W^T J^T.
\end{equation}

In addition to the center position $x$, each Gaussian is further defined by a scaling factor $s\in \mathbb{R}^{3}$, a rotation factor $r\in \mathbb{R}^{4}$, an opacity value $\alpha\in \mathbb{R}$, and spherical harmonic (SH) coefficients $c\in \mathbb{R}^{k}$ for colors with $k$ SH functions. For $N$ Gaussians, the complete 3D Gaussian representation is defined as
$\{x, s, r, \alpha, c\}$.

For dynamic scenes, the temporal deformation of 3D Gaussians, denoted as $\Delta \mathcal{G}$, is computed using a deformation field network $\mathcal{F}$. Given 3D Gaussians $\mathcal{G}$ and time $t$, the updated 3D Gaussians $\mathcal{G}^t$ are calculated as follows:

\begin{equation}
    \mathcal{G}^t = \mathcal{G} + \Delta\mathcal{G}, \quad \text{where} \quad \Delta\mathcal{G} = \mathcal{F}(\mathcal{G}, t).
\end{equation}

For a novel camera view with a view matrix $M$, the rendered scene at time $t$ can be formulated as:
\begin{equation}
        \hat{\mathcal{I}} = \mathcal{S}(M, \mathcal{F}(\mathcal{G}, t)+\mathcal{G}).
\end{equation}
where $\mathcal{S}$ denotes a differential splatting, which renders the Gaussians on 2D images.

In 4DGS~\cite{wu20244d}, the deformation field network consists of an encoder and a multi-head decoder. The encoder is constructed as a spatiotemporal structure encoder, \eg, HexPlane $\mathcal{H}$, and a small MLP $\phi_d$. As outlined in~\cite{cao2023hexplane}, the 3D Gaussian $\mathcal{G}$ and time $t$ are encoded as $\mathbf{f}_h$ by $\mathcal{H}$:
\begin{equation}
    \mathbf{f}_h = \mathcal{H}(\mathcal{G}, t),
\end{equation}
so that a 4D spacetime grid can be decomposed into six feature planes spanning each pair of coordinate axes.
$\mathbf{f}_h$ is further encoded as $\mathbf{f}_d$ by $\phi_d$, as follows:
\begin{equation}
    \mathbf{f}_d = \phi_{d}(\mathbf{f}_h).
\end{equation}
Then, the deformation output position shift $\Delta X$, scale change $\Delta s$, and rotation adjustment $\Delta r$ are generated by a Multi-head Gaussian Deformation Decoder $\mathcal{D}$, as follows:
\begin{equation}
   (\Delta x, \Delta s , \Delta r) = \mathcal{D}(\mathbf{f}_d).
\end{equation}

\subsection{Context awareness}
\label{sec:context}
In autonomous driving scenarios, scenes often change significantly over time, making it essential for the model to better understand the world during training. We aim to identify context-related information that can further optimize 4DGS’s learning process beyond RGB input. This approach has proven effective for 3DGS, as demonstrated in prior work in Feature 3DGS ~\cite{zhou2024feature}, which uses a 2D foundation model, \eg LSeg~\cite{li2022language}, to distill semantic-related features. Building on this, we extend the method to 4DGS by using a rasterizer that associates each Gaussian with a high-dimensional semantic feature during training, as depicted in Fig.~\ref{fig:system}. 

Specifically, when we obtain the semantic feature $\mathbf{f}_\textit{seg}$ from LSeg, we simultaneously generate a corresponding semantic feature deformation with each Gaussian deformation. The result of this deformation is then compared with the target feature in 2D, similar to the approach in 3DGS, using cosine similarity through a rasterizer.

Unlike Feature 3DGS, these features will deform along with the temporal deformation of the Gaussians, ensuring that the scene context remains consistent over time without the need for separate feature extraction for each static scene, as required in conventional static scene methods. We define these high dimensional semantic features as context awareness, \ie $\mathbf{f}_\textit{con} \leftarrow \mathbf{f}_\textit{seg}$.

\subsection{Temporal deformation awareness}
\label{sec:deformation}
A significant difference between 4DGS and 3DGS is that 4DGS requires time inputs to deform the Gaussians. In essence, this deformation is achieved by optimizing the 3D mapping through a 2D loss. When camera views are limited, however, 4DGS struggles to provide precise 3D Gaussian outputs at specific points in time. This limitation is particularly prominent in autonomous driving vision, where cameras typically follow a simple trajectory, such as a linear one. Given this scenario, we propose using deformations calculated based on a HexPlane as explicit inputs to an encoder, as follows:
\begin{equation}
        \mathbf{f}_{\textit{def}} \leftarrow \Delta \mathcal{G} = \mathcal{F}(\mathcal{G}, t),
\end{equation}
and together with other features to compensate for the temporal deformation.

\subsection{Time embedding}
\label{sec:time}

In addition, we also incorporate time information as a unique identifier for each frame. Specifically, we explicitly binarize the time step to obtain a sparse feature as follows:
\begin{equation}
       \mathbf{f}_\textit{time} = sin (\frac{\tau}{10000^{\frac{2i}{d}}}),
\end{equation}
where $\tau$ is the binary sparse feature of time step $t$, $i$ represents the index of the current dimension, and $d$ is the total dimension of the embedding. Each dimension of $\mathbf{f}_\text{time}$ captures time information with a different frequency, with lower dimensions being more sensitive to changes in $\tau$ and higher dimensions reacting more slowly. This design allows the embedding to capture multi-scale temporal patterns in the data. The sinusoidal function introduces periodicity, so even with very large or small values of $\tau$, the embedding remains within bounded ranges, enabling a consistent representation of time. 

\subsection{Deformation compensation network}
\label{sec:dcn}
We aim to refine the temporal deformation learned from time-based information, so we design a DCN to adjust the deformation inferred by the spatio-temporal structure encoder. The DCN consists of two channels: in the first channel, aggregated awareness features are fed into an MLP model to predict the potential deformation compensation needed for each Gaussian. This MLP is defined as ${\phi}_p$. The second channel comprises a linear layer and a set of sigmoid functions, denoted as ${\phi}_s$. Intuitively, this linear layer serves as a filter for the results learned by the first channel, enabling the DCN to refine the output with an additional layer of processing. 
We can represent the feature output from the DCN for Gaussian deformation compensation as follows:
\begin{equation}
       \mathcal{G}^t \leftarrow \mathcal{G}^t + {\phi}_p(\mathbf{f}_\textit{time}, \mathbf{f}_\textit{def},  \mathbf{f}_\textit{con}) \otimes {\phi}_s(\mathbf{f}_\textit{time}, \mathbf{f}_\textit{def},  \mathbf{f}_\textit{con}).
\end{equation}

The sigmoid-based filtering is designed because we observed that, during learning, certain local Gaussians, such as those representing the sky or ground, may undergo significant motion. While these Gaussians generally have minimal impact on rendering, they can introduce noise into the DCN. To mitigate this interference, we employ a sigmoid function to learn a mask that filters out disturbances caused by irrelevant Gaussian motions, thereby limiting their influence during the compensation phase. Ultimately, these additional deformation adjustments are applied to refine the results from the HexPlane stage.

\subsection{Optimization}
\label{sec:opt}
\mypara{Initialization}.
\cite{yan2024street} suggests using LiDAR points from the ego vehicle for initialization rather than relying on structure from Motion (SfM)~\cite{schonberger2016structure}. The dense point cloud from LiDAR provides a robust initialization, allowing the 3D Gaussian to capture unseen areas during driving. We initialize the 4D model based on a pre-trained 3D Gaussian, which is initialized using LiDAR points from all frames.

\mypara{Loss}.
We use an L1 loss $\mathcal{L}_{\text{rgb}}$ to supervise the rendered RGB images, along with a grid-based total variation loss and a D-SSIM term $\mathcal{L}_{\textit{d-ssim}}$ to enhance structural similarity~\cite{wang2004image}. Following~\cite{wu20244d}, we consider grid-based total-variational loss $\mathcal{L}_{\textit{tv}}$ into the optimization, with weight $\lambda_{\textit{tv}}$
As recommended in~\cite{huang2024textit}, we also include a depth loss $\mathcal{L}_{\textit{depth}}$, and following \cite{zhou2024feature}, we incorporate a feature loss $\mathcal{L}_f$ with a weighting factor $\lambda_f$, the loss is formulated as follows:
\begin{multline}
        \mathcal{L} = \lambda_{\textit{rgb}} \mathcal{L}_\textit{rgb} +  \lambda_{\textit{d-ssim}} \mathcal{L}_{\textit{d-ssim}} + \lambda_\textit{tv} \mathcal{L}_\textit{tv} \\ + \lambda_\textit{depth} \mathcal{L}_\textit{depth} + \lambda_f \mathcal{L}_\textit{f}.
\end{multline}

%% file: sec/4_experiments.tex
\setlength{\tabcolsep}{0.2pt}
\renewcommand{\arraystretch}{0.8} 
\setlength{\tabcolsep}{3pt}
\begin{table}[t!]
\centering
\fontsize{7}{12}\selectfont
\begin{threeparttable}
\caption{4D reconstruction and novel view synthesis performance comparison on the Waymo Open dataset~\cite{sun2020scalability}.}
\label{table:comparison_waymo}
\begin{tabular}{L{2cm}|C{0.8cm}C{0.8cm}C{0.8cm}|C{0.8cm}C{0.8cm}C{0.8cm}}
\toprule 
\centering{Tasks} & \multicolumn{3}{c|}{4D Reconstruction} & \multicolumn{3}{c}{Noval View Synthesis} \\
\midrule 
\centering{Metrics} & PSNR↑  & SSIM↑ & LPIPS↓ & PSNR↑  & SSIM↑ & LPIPS↓\\
\midrule 
StreetSurf~\cite{guo2023streetsurf} & 25.32 & 0.764 & 0.377 & 23.67 & 0.672 & 0.411 \\
NSG~\cite{ost2021neural} & 24.11 & 0.654 &  0.401 & 21.05 & 0.579 & 0.477\\
MARS~\cite{wu2023mars} & 26.51 & 0.792 &  0.339 & 23.14 & 0.716 & 0.403\\
SUDS~\cite{turki2023suds} & 27.74 & 0.822 & 0.221 & 21.17 & 0.638 & 0.442\\
3DGS~\cite{kerbl3Dgaussians} & 26.97 & 0.897 & 0.229 & 25.11 & 0.827  & 0.298\\
EmerNeRF~\cite{yang2023emernerf} & 27.07 & 0.841 & 0.317 & 25.28 & 0.746 & 0.309 \\
4DGS~\cite{liu2023real} & 31.02  & 0.901 & 0.136 & 26.41 & 0.814 & 0.191\\
$S^3$Gaussian~\cite{huang2024textit} & 32.16  & 0.915 & 0.101  & 27.51 & 0.862 & 0.100\\
\textbf{CoDa-4DGS (ours)} & \textbf{33.65} & \textbf{0.919} & \textbf{0.078} &  \textbf{28.66}& \textbf{0.900} & \textbf{0.058}\\
\bottomrule 
\end{tabular}
\end{threeparttable}
\end{table}

\setlength{\tabcolsep}{0.2pt}
\renewcommand{\arraystretch}{0.8} 
\setlength{\tabcolsep}{3pt}
\begin{table}[t!]
\centering
\fontsize{7}{12}\selectfont
\begin{threeparttable}
\caption{4D reconstruction and novel view synthesis performance comparison on the KITTI dataset~\cite{geiger2013vision}.}
\label{table:comparison_kitti}
\begin{tabular}{L{2cm}|C{0.8cm}C{0.8cm}C{0.8cm}|C{0.8cm}C{0.8cm}C{0.8cm}}
\toprule 
\centering{Tasks} & \multicolumn{3}{c|}{4D Reconstruction} & \multicolumn{3}{c}{Noval View Synthesis} \\
\midrule 
\centering{Metrics} & PSNR↑  & SSIM↑ & LPIPS↓ & PSNR↑  & SSIM↑ & LPIPS↓\\
\midrule 
StreetSurf~\cite{guo2023streetsurf} & 24.25 & 0.821 & 0.251 & 22.47 & 0.756 & 0.312 \\
NSG~\cite{ost2021neural} & 26.66 & 0.806 &  0.186 & 21.53 & 0.673 & 0.254\\
MARS~\cite{wu2023mars} & 27.96 & 0.900 &  0.185 & 24.23 & 0.845 & 0.160\\
SUDS~\cite{turki2023suds} & 28.31 & 0.876 & 0.185 & 22.77 & 0.797 & 0.171\\
3DGS~\cite{kerbl3Dgaussians} & 20.72 & 0.774 & 0.222&19.14& 0.736 &0.237\\
EmerNeRF~\cite{yang2023emernerf} & 26.89 & 0.819 & 0.221 & 24.99 & 0.791 & 0.245 \\
4DGS~\cite{liu2023real}  & 29.99 & 0.881 & 0.143 & 25.84 & 0.839 & 0.192\\
$S^3$Gaussian~\cite{huang2024textit} & 31.07 & 0.899 & 0.101 & 26.14 & 0.852 & 0.146\\
\textbf{CoDa-4DGS (ours)} & \textbf{32.31} & \textbf{0.907} & \textbf{0.083} & \textbf{27.16} & \textbf{0.897} & \textbf{0.118}\\
\bottomrule 
\end{tabular}
\end{threeparttable}
\end{table}

\setlength{\tabcolsep}{3pt}
\begin{table}[t!]
\centering
\fontsize{8}{12}\selectfont
\begin{threeparttable}
\caption{4D Reconstruction Performance Comparison on Dynamic-32 in the NOTR Benchmark~\cite{yang2023emernerf}.}
\label{table:comparison1}
\begin{tabular}{L{2.3cm}|C{0.9cm}C{0.9cm}C{0.9cm}C{0.9cm}C{0.9cm}}
\toprule 
\centering{Metrics} & PSNR↑   & SSIM↑ & LPIPS↓ & PSNR*↑   & SSIM*↑  \\
\midrule 
3DGS~\cite{kerbl3Dgaussians} & 26.50 & 0.897 & 0.229 & 21.43 & 0.639 \\
EmerNeRF~\cite{yang2023emernerf} & 28.16 & 0.806  & 0.228 & 24.32 & 0.682\\
4DGS~\cite{liu2023real} & 31.04 &	0.908 &	0.124	& 26.88	& 0.831  \\
$S^3$Gaussian~\cite{huang2024textit} & 32.39 & 0.920
 & 0.097 & 28.39 & 0.850  \\
\textbf{\myMethod (ours)}  & \textbf{32.86}
 & \textbf{0.927}  & \textbf{0.086}  & \textbf{29.25} &  \textbf{0.866}  \\
\bottomrule 
\end{tabular}
\begin{tablenotes}
    \scriptsize 
    \item[] * indicates that the metrics are calculated only for dynamic objects.
\end{tablenotes}
\end{threeparttable}
\end{table}

\setlength{\tabcolsep}{3pt}
\begin{table}[t!]
\centering
\fontsize{8}{12}\selectfont
\begin{threeparttable}
\caption{Ablation study on the Deformation Compensation Network (DCN).}
\label{table:ablation2}
\begin{tabular}{L{1.9cm}|C{0.7cm}|C{0.8cm}C{0.8cm}C{0.8cm}C{0.9cm}C{0.9cm}}
\toprule 
\centering{Metrics}  & \# Par. & PSNR↑   & SSIM↑ & LPIPS↓   & PSNR*↑ & SSIM*↑ \\
\midrule 
w/o. DCN &35.8M & 31.71 & 0.911 & 0.091 & 28.05  & 0.850 \\
w. deeper MLP   &36.6M  & 31.75 & 0.911  & 0.090  &28.08 & 0.853 \\
w. DCN  & 36.4M& \textbf{32.98} & \textbf{0.929} & \textbf{0.078} & \textbf{28.75} & \textbf{0.867} \\

\bottomrule 
\end{tabular}
\begin{tablenotes}
    \scriptsize 
    \item[] * indicates that the metrics are calculated only for dynamic objects.
\end{tablenotes}
\end{threeparttable}
\end{table}

\begin{figure*}[t!]
   \centering
   \includegraphics[width=1\textwidth]{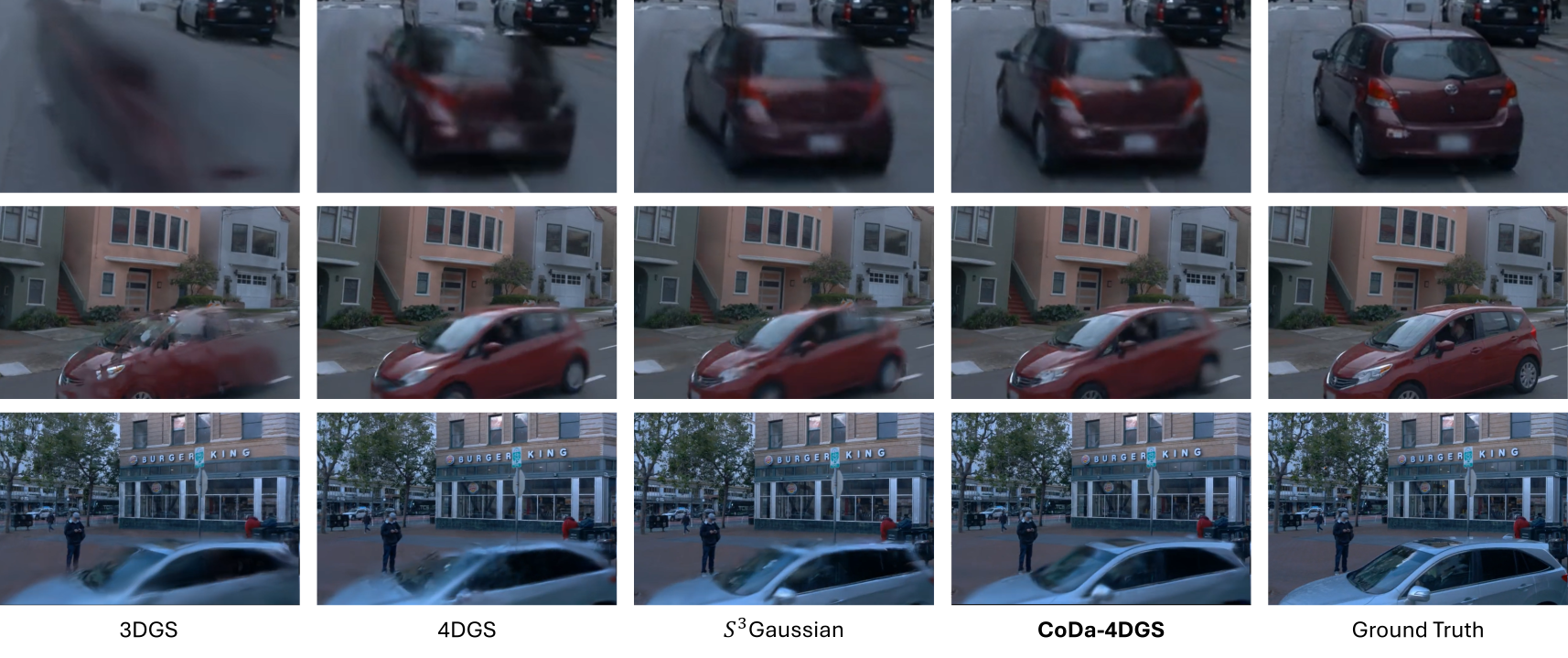}
   \caption{Visual comparison of rendering results from \myMethod and other approaches on Waymo Open dataset.}
   \label{fig:visual_reswaymo}
\end{figure*}

\begin{figure}[t!]
   \centering
   \includegraphics[trim={7cm 4.5cm 5cm 2.8cm},clip, width=0.48\textwidth]{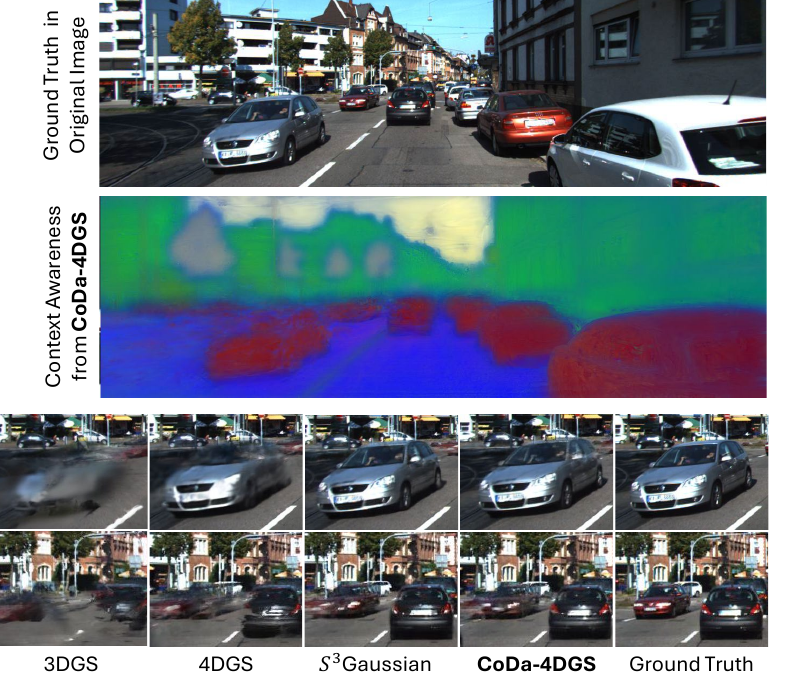}
   \caption{Visual comparison of rendering results from CoDa-4DGS and other approaches on KITTI dataset.}
   \label{fig:visual_reskitti}
\end{figure}

\setlength{\tabcolsep}{3pt}
\begin{table}[t!]
\centering
\fontsize{8}{12}\selectfont
\begin{threeparttable}
\caption{Ablation study on time embedding ($\mathbf{f}_\textit{time}$), context awareness ($\mathbf{f}_\textit{con}$) and temporal deformation awareness ($\mathbf{f}_\textit{def}$).}
\label{table:ablation1}
\begin{tabular}{C{0.7cm}C{0.7cm}C{0.7cm}|C{0.9cm}C{0.9cm}C{0.9cm}C{0.9cm}C{0.9cm}}
\toprule 
$\mathbf{f}_\textit{time}$ & $\mathbf{f}_\textit{con}$ & $\mathbf{f}_\textit{def}$  & PSNR↑   & SSIM↑ & LPIPS↓ &PSNR*↑ & SSIM*↑ \\
\midrule 
\checkmark  & \checkmark &  & 32.80&   0.925 &0.088   & 29.21&0.863  \\
\checkmark  &  & \checkmark &  32.77 & 0.924   &0.089 & 29.19&0.863 \\
 & \checkmark & \checkmark &  32.69   &  0.924 & 0.093  & 29.12&0.862\\
\checkmark & \checkmark & \checkmark &  \textbf{32.86} & \textbf{0.927} & \textbf{0.086}& \textbf{29.25}& \textbf{0.866}\\
\bottomrule 
\end{tabular}
\begin{tablenotes}
    \scriptsize 
    \item[] * indicates that the metrics are calculated only for dynamic objects.
\end{tablenotes}
\end{threeparttable}
\end{table}
\section{Experimental evaluation}
\label{sec:experiments}
In this section, we evaluate CoDa-4DGS against other baselines quantitatively and qualitatively. We conduct ablation studies on each module of CoDa-4DGS and showcase its potential applications.

\subsection{Experimental setup}
\label{sec:setup}

We use the Waymo Open dataset~\cite{sun2020scalability}, KITTI dataset~\cite{geiger2013vision} and NeRF On-The-Road (NOTR) dataset~\cite{yang2023emernerf} as our primary benchmarks. We evaluate our approach on two key tasks: 4D reconstruction and novel view synthesis. The main evaluation metrics include Peak Signal-to-Noise Ratio (PSNR), Structural Similarity Index (SSIM)~\cite{wang2004image}, and Learned Perceptual Image Patch Similarity (LPIPS)~\cite{dosovitskiy2016generating}. Our baselines set of a series of methods based on NeRF and Gaussian splatting, including StreetSurf~\cite{guo2023streetsurf}, NSG~\cite{ost2021neural}, MARS~\cite{wu2023mars}, SUDS~\cite{turki2023suds}, EmerNeRF~\cite{yang2023emernerf}, 3DGS~\cite{kerbl3Dgaussians}, 4DGS~\cite{wu20244d} and $S^3$Gaussian~\cite{huang2024textit}. More details of the implementation can be found in Sec.~\ref{sec:imp_details}.

\subsection{Comparison}
\label{sec:comparison}

\noindent\textbf{Quantitatively evaluation}.
Following previous works~\cite{huang2024textit,wu2023mars}, we evaluate CoDa-4DGS and other baselines for both 4D reconstruction and novel view synthesis on Waymo and KITTI datasets. The results in Tab.~\ref{table:comparison_waymo} and Tab.~\ref{table:comparison_kitti} show that CoDa-4DGS outperforms other baselines across all metrics. To further validate 4D rendering performance on dynamic objects, we evaluate all methods on the NOTR Dynamic-32 benchmark suggested in~\cite{yang2023emernerf}, which focuses on dynamic scene evaluation. Alongside PSNR, SSIM, and LPIPS, we also evaluate the rendering quality specifically for dynamic objects, denoted as PSNR* and SSIM*. As shown in Tab.~\ref{table:comparison1}, CoDa-4DGS performs better than all baselines.

\begin{figure}[t!]
   \centering
   \includegraphics[trim={0cm 0cm 0cm 0cm},clip, width=0.48\textwidth]{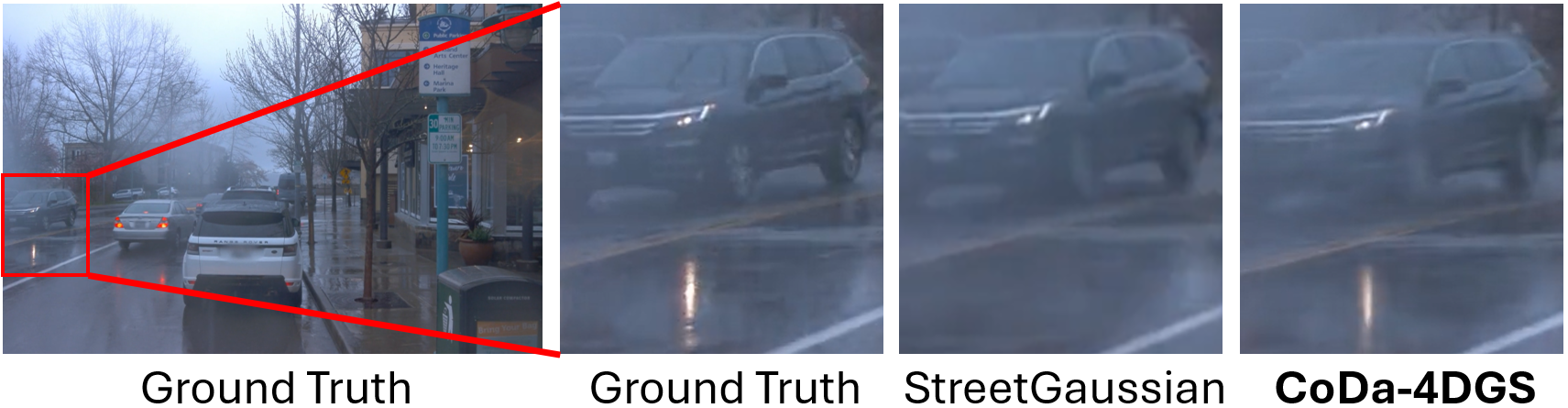}
   \caption{Visual comparison of novel view synthesis between CoDa-4DGS and StreetGaussian~\cite{yan2024street}. Due to its independence from bounding box labels for dynamic objects, CoDa-4DGS can surpass StreetGaussian in rendering dynamic details outside these bounding boxes (moving light reflections on wet roads).}
   \label{fig:comp_streetgs}
\end{figure}

\begin{figure}[t!]
   \centering
   \includegraphics[width=0.48\textwidth]{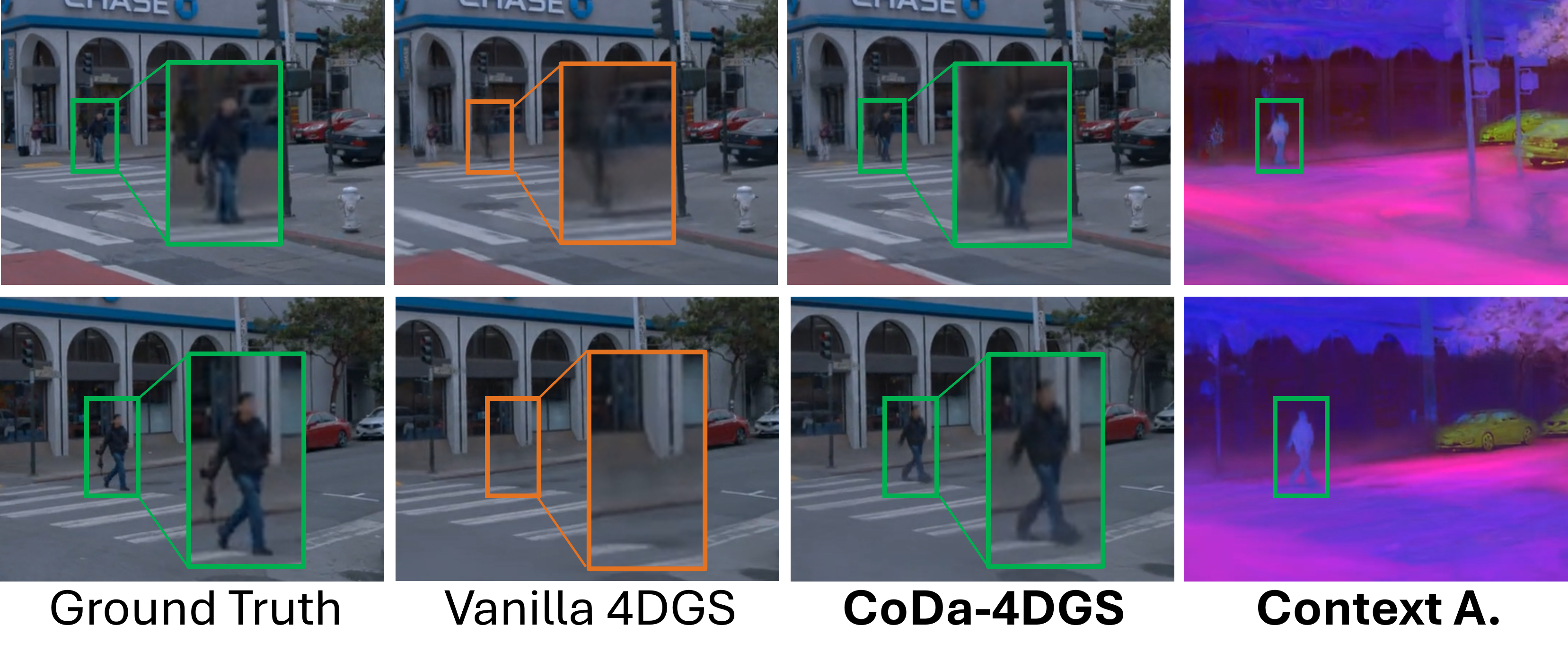}
   \caption{Visual comparison of novel view synthesis between vanilla 4DGS~\cite{liu2023real} and CoDa-4DGS. Due to the integration of context awareness (context A.), CoDa-4DGS can accurately render a pedestrian walking out from the shadows (from time $t$ to $t+\Delta t$), achieving better results than the vanilla 4DGS~\cite{wu20244d}.}
   \label{fig:comp_4DGS}
\end{figure}

\noindent\textbf{Qualitatively analysis}.
In Fig.~\ref{fig:visual_reswaymo} and Fig.~\ref{fig:visual_reskitti}, we present a visual comparison between CoDa-4DGS and the baselines. The results show that CoDa-4DGS has a clear advantage in rendering fine details and contours of moving objects. Notably, compared to supervised dynamic object rendering methods that rely on bounding box annotations, such as StreetGaussian~\cite{yan2024street}, CoDa-4DGS, as a bounding-box-free self-supervised approach, exhibits distinct strengths and trade-offs in dynamic object rendering: (\emph{i}) Self-supervised dynamic rendering offers better scalability since it eliminates the need for manual annotations; (\emph{ii}) Bounding-box-based methods may overlook certain dynamic elements that extend beyond bounding boxes. Fig.~\ref{fig:comp_streetgs} shows that CoDa-4DGS successfully renders dynamic details outside bounding boxes, such as the moving reflections of car headlights on a wet road, whereas StreetGaussian fails to capture this effect. Furthermore, we compare CoDa-4DGS with vanilla 4DGS on pedestrian rendering in Fig.~\ref{fig:comp_4DGS}. The scene includes a pedestrian walking from the shadows, where 4DGS struggles to render the pedestrian consistently over time. However, due to the inputs from context awareness, CoDa-4DGS effectively maintains the rendering of the moving pedestrian across frames, showcasing its superior temporal coherence.

\subsection{Ablation study}
\label{sec:ablation}
\begin{figure}[t!]
   \centering
   \includegraphics[width=0.48\textwidth]{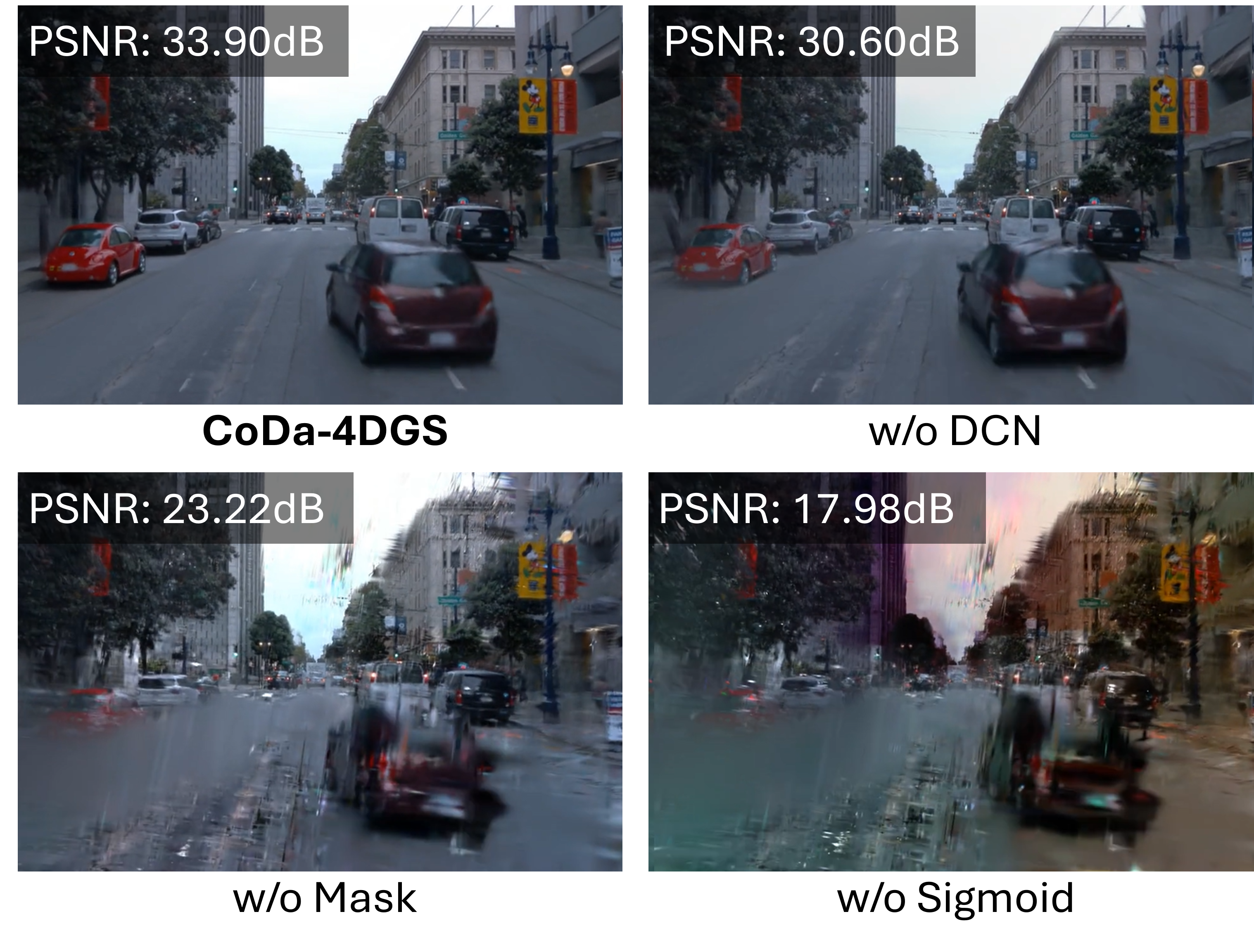}
   \caption{Visual ablation study on Deformation Compensation Network (DCN).}
   \label{fig:ablation_dcn}
\end{figure}

\begin{figure}[t!]
   \centering
   \includegraphics[width=0.48\textwidth]{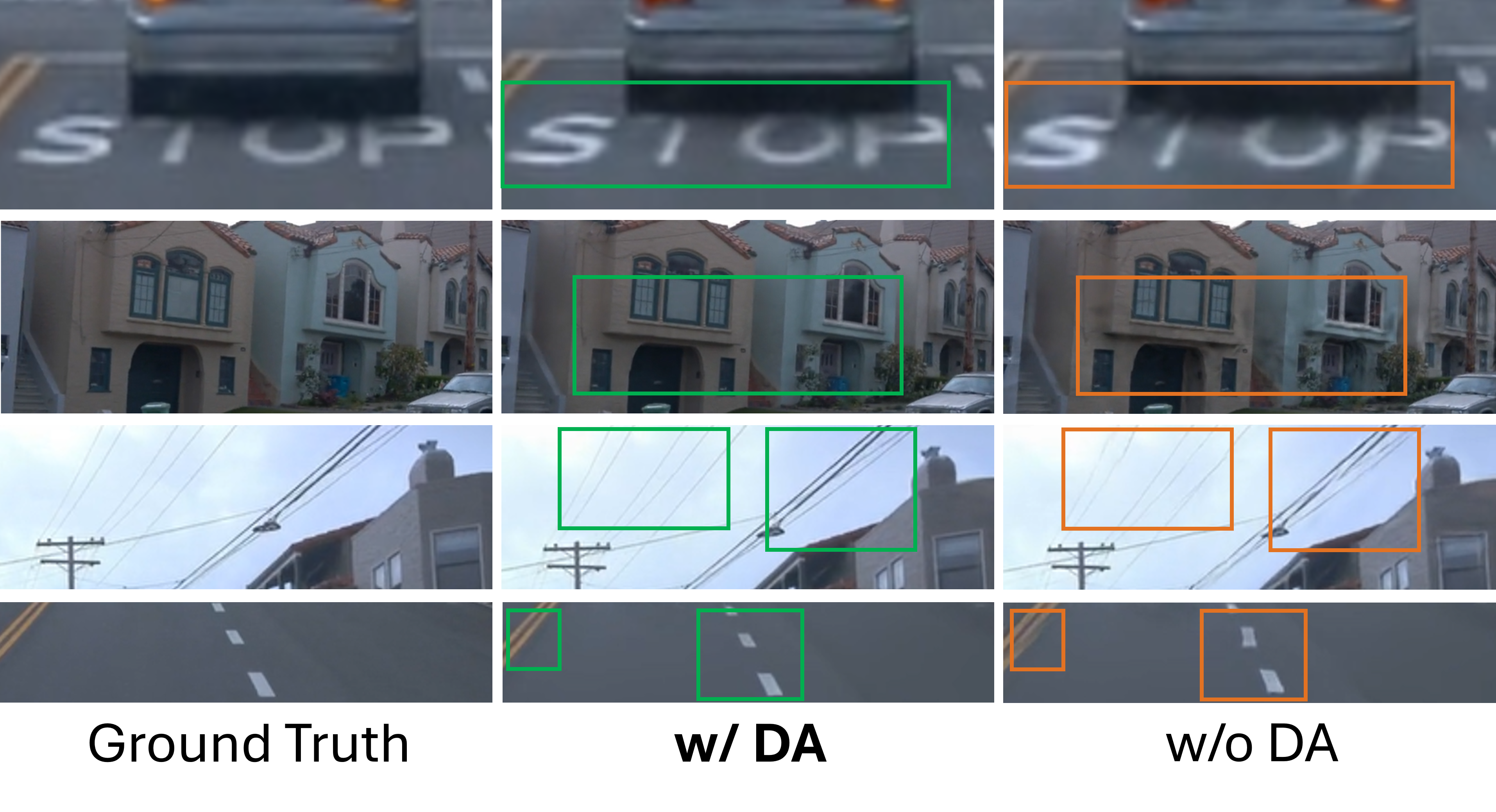}
   \caption{Visual ablation study on Deformation Awareness (DA).}
   \label{fig:ablation_da}
\end{figure}

\noindent\textbf{Refinement through DCN}.
We first validate the effectiveness of DCN in Tab.~\ref{table:ablation1}, where removing DCN results in decreased rendering performance across all metrics. To ensure a fair comparison that mitigates any effects from increased model size, the depth of the original MLP is increased to match the parameter count of our method when DCN is removed. The results demonstrate that simply deepening the MLP does not significantly enhance rendering quality. Furthermore, we visualize the impact of DCN in Fig.~\ref{fig:ablation_dcn} showing that DCN enhances the clarity of rendering results for dynamic objects, particularly by sharpening their contours. We also evaluate the effects of the mask and signmod components within DCN. The experimental results in Fig.~\ref{fig:ablation_dcn} show that the Sigmoid function helps filter noise in the latent space, while the DCN mask minimizes the impact of unnecessary Gaussian movements in areas of similar color. Without the mask, DCN leads to blurred roads and unnatural cracks in the sky. These findings align with the original design principles of DCN.

\noindent\textbf{Contribution of awareness}.
We evaluate the roles of context awareness, deformation awareness, and time embedding in our model. As shown in Tab.~\ref{table:ablation2}, removing any one of these inputs leads to a decrease in performance on all metrics. Beyond quantitative evaluation, we also conduct a visual analysis to examine how context awareness and deformation awareness affect the rendered results. As shown in Fig.~\ref{fig:comp_4DGS}, context awareness helps CoDa-4DGS identify dynamic objects that are difficult to distinguish in RGB images, such as pedestrians emerging from shadows. Furthermore, in Fig.~\ref{fig:ablation_da}, a comparison between rendering with and without deformation awareness reveals its contribution to improving fine details in the rendered scenes. Note that while these improvements from deformation awareness may not be strongly reflected in quantitative enhancement, they can enhance the visual realism of the rendered results.

\subsection{Applications}
\label{sec:app}
By leveraging semantic features as context awareness and applying distillation, these semantic attributes undergo temporal deformation along with the Gaussians. As a result, each Gaussian preserves the continuity of its semantic features over time, enabling a wider range of applications. Fig.~\ref{fig:teaser} shows some examples of these applications.

In 4D reconstruction and novel view synthesis applications, CoDa-4DGS not only supports RGB-based rendering like other 4DGS-based methods in~\cite{liu2023real,huang2024textit}, but also extends this task to semantic understanding using foundation models. Moreover, by performing semantic segmentation on Gaussians, CoDa-4DGS enables prompt-based 4D semantic segmentation in the RGB domain, which can further be applied to 2D pixel-level semantic segmentation. Unlike previous work in~\cite{yan2024street,wu2023mars}, CoDa-4DGS also introduces bounding-box-free instance decomposition through feature extraction. This allows extracted instances to be synthesized and integrated into new scenes. By manipulating the pose and position of both existing and newly synthesized objects, CoDa-4DGS unlocks new possibilities for flexible 4D scene editing.

%% file: sec/5_conclusion.tex
\section{Conclusion}
\label{sec:conclusion}
In this work, we explore enhancing 4DGS for dynamic scene rendering in autonomous driving by learning deformation compensation. We introduce \myMethod, which integrates context and deformation awareness along with temporal encoding into a DCN module. This setup learns and filters features to compensate for the deformations captured from vanilla 4DGS, thereby further optimizing dynamic rendering outcomes. Validation on autonomous driving datasets shows that our method can effectively enhance vanilla 4DGS, with particularly notable improvements in rendering complex scenes and dynamic objects.

\noindent\textbf{Limitations}. While \myMethod can be directly integrated into Vanilla 4DGS, the inclusion of the DCN and the rasterizer used for distilling semantic features increases the model's parameter count. 
Given an $n$-dimensional feature for each Gaussian, the computational complexity of CoDa-4DGS is 
$\mathcal{O}(n)$. Incorporating a 128-dimensional awareness feature expands each Gaussian's feature dimension from 62 to 190, resulting in a $2.1\times$ increase in computational complexity, which aligns with our measurement results. Regarding memory requirements, assuming 1 million Gaussians during training, using a 128-dimensional awareness feature requires approximately 4.1 GB of memory.

%% file: sec/x1_imp_details.tex
\section{Implementation details}
\label{sec:imp_details}

In the implementation, for context awareness, we follow and use LSeg~\cite{li2022language} to maintain 128-dimensional semantic features that link each Gaussian with temporal deformation. Thus, context awareness is represented by aggregated semantic features across all Gaussians, \ie $\mathbf{f}_{\textit{con}} \in \mathbb{R}^{N\times 128}$. Temporal deformation awareness is built on $\Delta \mathcal{G}$, such that $\mathbf{f}_{\textit{def}} \in \mathbb{R}^{N\times 62}$, where for SH coefficients $k = 48$. Additionally, the frame information is binarized and encoded as a periodic function to generate a time embedding $\mathbf{f}_{\textit{time}} \in \mathbb{R}^{N\times 64}$. Since our primary comparison is with $S^3$Gaussian~\cite{huang2024textit}, we adopted similar hyperparameters. We train for 50,000 steps, with a learning rate set to $1.6e^{-3}$, which decays to $1.6e^{-4}$. For our loss function, we assign  weights for each as follows:
$ \lambda_{\textit{rgb}} = 1$,
$ \lambda_{\textit{d-ssim}}= 0.2$,
$ \lambda_\textit{tv} =1 $,
$ \lambda_\textit{depth} = 0.5$,
$ \lambda_f  = 1$.
        

%% file: sec/x2_plug_and_play.tex
\section{Plug-and-play}
\label{sec:plugplay}

\setlength{\tabcolsep}{3pt}
\begin{table*}[t!]
\centering
\fontsize{9}{12}\selectfont
\begin{threeparttable}
\caption{Enhancing scene reconstruction accuracy via Plug-and-Play integration of CoDa-4DGS for Scene 22 and Scene 02.}
\label{table:plug-in}
\begin{tabular}{L{4.0cm}||C{1.2cm}C{1.2cm}C{1.2cm}C{1.2cm}||C{1.2cm}C{1.2cm}C{1.2cm}C{1.2cm}}
\toprule 
\centering{Method} & \multicolumn{4}{c||}{Scene 22} & \multicolumn{4}{c}{Scene 02} \\
\midrule 
\centering{Metric} & PSNR↑   & SSIM↑   & PSNR*↑ & SSIM*↑ & PSNR↑   & SSIM↑   & PSNR*↑ & SSIM*↑ \\
\midrule 
4DGS~\papersource{(CVPR~24)} & 32.54 & 0.9327 & 29.77 & 0.8767 & 29.14 & 0.8837 & 24.19 & 0.7965 \\
4DGS+CoDa-4DGS & 33.12 & 0.9437 & 30.53 & 0.8802 & 29.84 & 0.8902 & 24.96 & 0.8042 \\
\cellcolor{gray!20} Improvement & \cellcolor{gray!20}+0.58 & \cellcolor{gray!20}+0.0110 & \cellcolor{gray!20}+0.76 & \cellcolor{gray!20}+0.0035 & \cellcolor{gray!20}+0.70 & \cellcolor{gray!20}+0.0065 & \cellcolor{gray!20}+0.77 & \cellcolor{gray!20}+0.0077 \\
\midrule 
$S^3$Gaussian~\papersource{(ECCV~24)} & 33.49 & 0.9367 & 29.79 & 0.8832 & 30.14 & 0.8998 & 24.78 & 0.8087 \\
$S^3$Gaussian+CoDa-4DGS & 34.09 & 0.9441 & 31.14 & 0.8935 & 30.37 & 0.9030 & 25.26 & 0.8150 \\
\cellcolor{gray!20} Improvement & \cellcolor{gray!20}+0.60 & \cellcolor{gray!20}+0.0074 & \cellcolor{gray!20}+1.35 & \cellcolor{gray!20}+0.0103 & \cellcolor{gray!20}+0.23 & \cellcolor{gray!20}+0.0032 & \cellcolor{gray!20}+0.48 & \cellcolor{gray!20}+0.0063 \\
\bottomrule 
\end{tabular}
\begin{tablenotes}
    \scriptsize 
    \item[1] *  indicates that the metrics are calculated only for dynamic objects.
\end{tablenotes}
\end{threeparttable}
\end{table*}

The core functionality of CoDa-4DGS lies in extracting temporal deformation awareness and context awareness, followed by Gaussian deformation compensation using DCN. This streamlined interface design makes CoDa-4DGS a plug-and-play method. When integrating CoDa-4DGS, we only need to focus on two aspects: acquiring temporal deformation awareness and context awareness. Since context awareness depends on components related to 2D foundation models, selecting an appropriate foundation model is essential to complement CoDa-4DGS effectively. For temporal deformation awareness, it is crucial to ensure that the embedded method can extract temporal deformation, such as the vanilla 4DGS.

To validate the plug-and-play functionality of CoDa-4DGS and its performance improvements over baseline methods, we conducted ablation studies using vanilla 4DGS and $S^3$Gaussian on Scene 22 and Scene 02, respectively. To ensure a fair comparison, we used identical hyperparameters, including learning rate, number of iterations, and the number of frames. As shown in Tab.~\ref{table:plug-in}, incorporating CoDa-4DGS led to performance improvements across all metrics for both vanilla 4DGS and $S^3$Gaussian, with approximately a 2\% increase in global PSNR. Notably, the improvement in dynamic PSNR was even more significant, aligning with the findings in the main paper. These results demonstrate that CoDa-4DGS effectively enhances rendering performance for dynamic objects through deformation compensation, making it particularly beneficial for autonomous driving scenarios.

%% file: sec/x3_scenario_editor.tex
\section{Scene editing}
\label{sec:scene_editing}

\begin{figure*}[t!]
   \centering
   \includegraphics[width=1\textwidth]{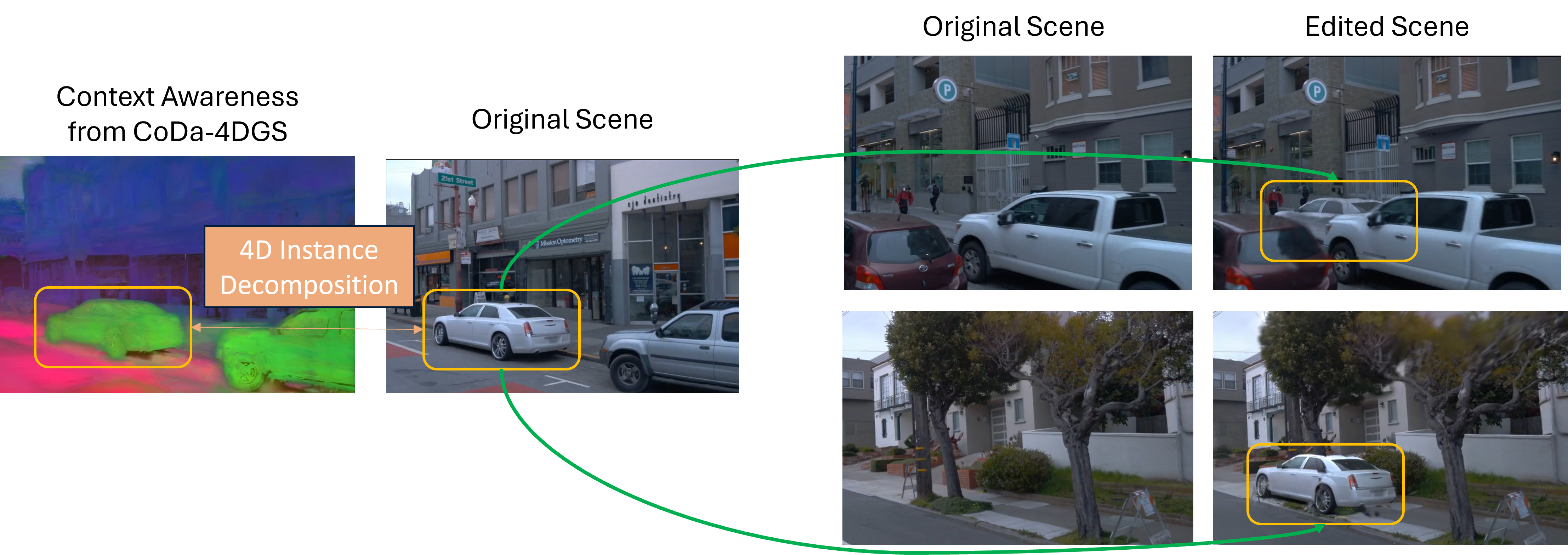}
   \caption{Object decomposition for scene editing. We extract an object from one scene (Scene 22) and integrate it into other scenes (Top: Scene 16 and Bottom: Scene 86). This process involves fusing two Gaussian models, leveraging instance decomposition in a 4D spatial-temporal space to achieve realistic synthesis and consistent object placement.}
   \label{fig:displacement}
\end{figure*}

\begin{figure*}[t!]
   \centering
   \includegraphics[width=1\textwidth]{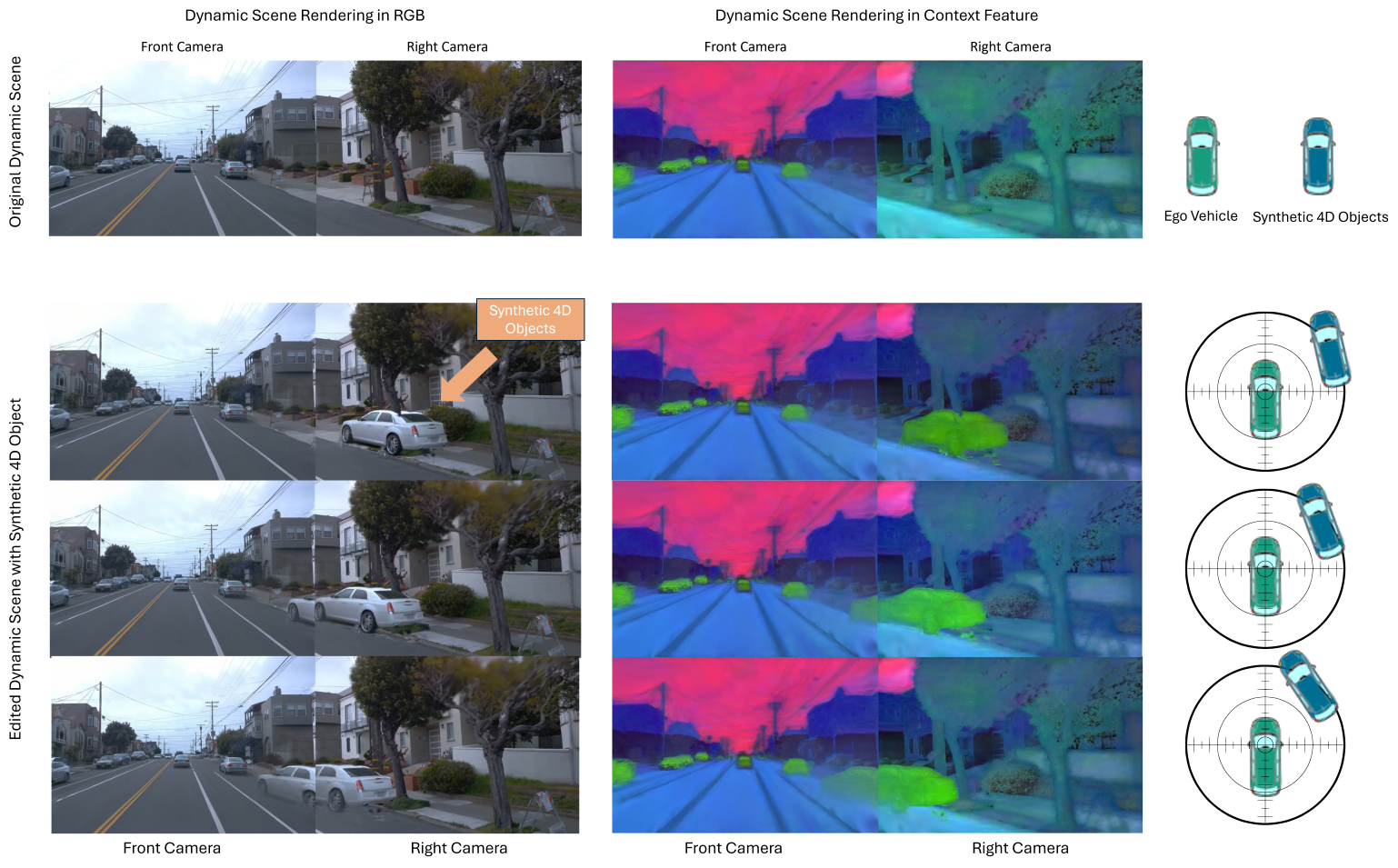}
   \caption{Manipulating synthetic 4D objects for scene editing by adjusting their poses and positions. This is achieved through translation and rotation of the object's Gaussian representations.}
   \label{fig:4dssyntheticobj}
\end{figure*}

In CoDa-4DGS, each Gaussian is trained to encode 4D semantic features, enabling context awareness. This allows us to use text encoding to generate a corresponding reference feature and, through cosine similarity and clustering, decompose the associated 4D Gaussians for a specific instance.

In Fig.~\ref{fig:displacement}, we demonstrate how context awareness can be used to extract an object from one scene and place it into another. Notably, the rendering of the edited scene is not achieved by directly rasterizing the object's corresponding 4D Gaussian onto the previous image. Instead, the 4D Gaussians are merged, allowing the object's Gaussian to deform in tandem with the temporal dynamics of the new scene. As shown in Figure 1, the spatial relationships inherent in the 4D Gaussians ensure that the newly added object can be partially occluded by other vehicles in the scene.

In Fig.~\ref{fig:4dssyntheticobj}, we showcase how the newly added synthetic 4D object can be manipulated within the new scene. By simultaneously rotating and translating the vehicle, the synthetic 4D object can be positioned with various poses in different locations. In the attached video, we provide demonstrations of these capabilities.

%% file: sec/x4_novel_view_synthesis.tex
\section{Novel view synthesis}
\label{sec:nvs}

\begin{figure*}[t!]
   \centering
   \includegraphics[width=1\textwidth]{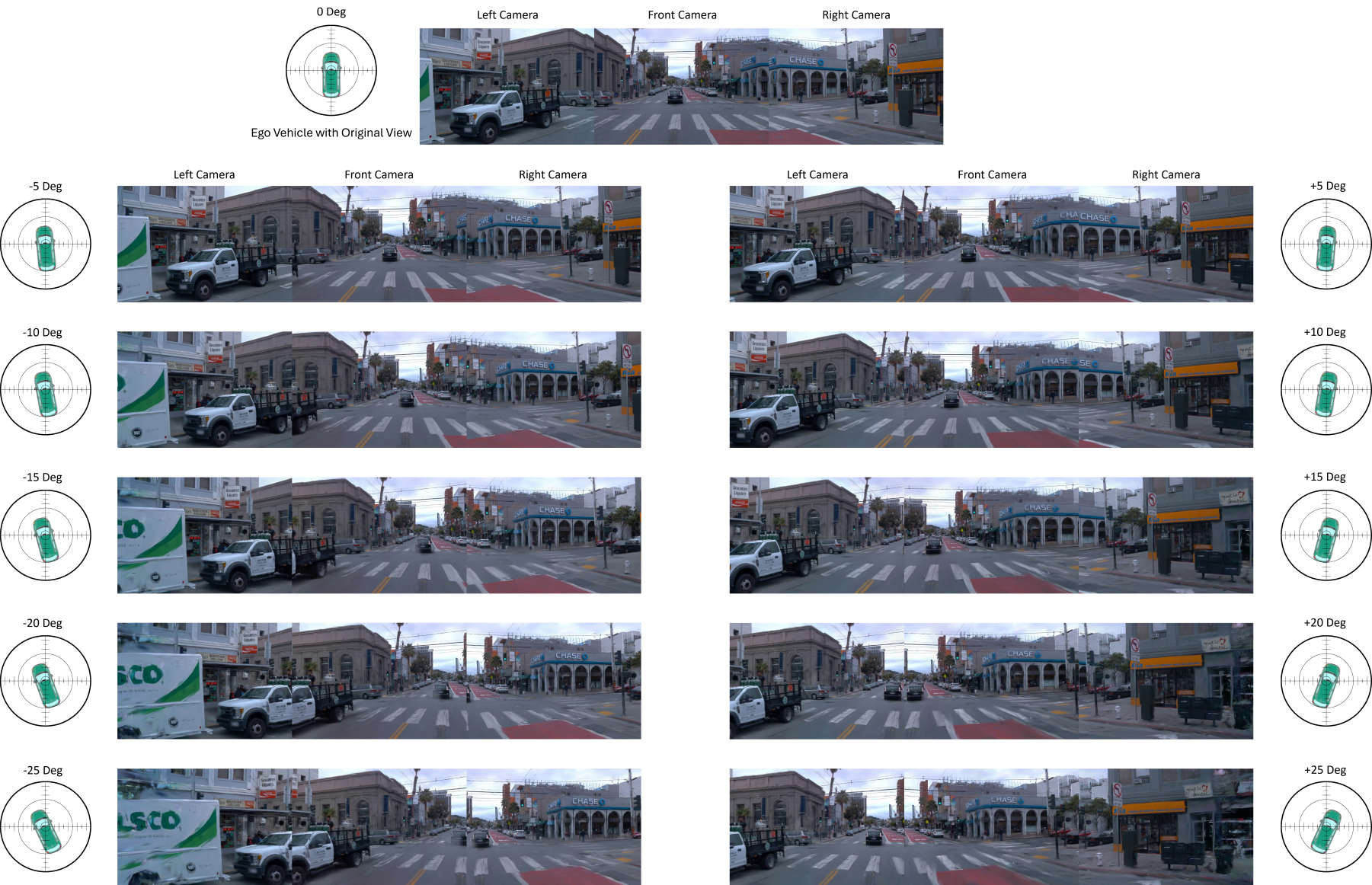}
   \caption{Novel view synthesis by freezing the scene in time and altering the cameras' perspective. The resulting views are synthetic images that do not exist in the dataset and have never been observed before.}
   \label{fig:nvs}
\end{figure*}

Novel view synthesis involves rendering camera perspectives that were not included in the training data. In autonomous driving, this capability is essential for photorealistic closed-loop simulations, particularly in validating end-to-end autonomous driving systems. Current benchmarks address this challenge primarily in two ways: (\emph{i}) utilizing simulator engines like CARLA~\cite{dosovitskiy2017carla}, and (\emph{ii}) leveraging Bird's Eye View (BEV) abstractions, as exemplified by the method proposed in NAVSIM~\cite{Dauner2024NEURIPS}, which serves as a benchmark for the CVPR 2024 Autonomous Grand Challenge. NAVSIM enables short closed-loop simulations built on the nuPlan dataset~\cite{caesar2021nuplan}. However, achieving photorealistic novel view synthesis remains a significant challenge.

This difficulty stems from the nature of real-world data collected for autonomous driving, which is typically captured using cameras mounted on vehicles. Camera movements are constrained by vehicle trajectories, often limited to simple, linear paths or curves. These restrictions in the training data result in scene reconstructions that struggle to meet the requirements for novel view synthesis.

In studies such as EmerNeRF~\cite{yang2023emernerf}, MARS~\cite{wu2023mars}, and StreetGaussian~\cite{yan2024street}, novel view synthesis is evaluated by dividing a scene's frames into training and test sets. For instance, EmerNeRF designates every 10th frame as the test set and the rest as the train set. While this benchmark method provides ground truth for benchmarking, and we adopt the same approach for quantitatively evaluating novel view synthesis performance, it falls short in meeting the requirements of closed-loop simulation, where novel views must be generated under diverse ego poses and dynamic trajectories.

Toward photorealistic closed-loop simulation, we showcase the capabilities of CoDa-4DGS in Fig.~\ref{fig:nvs}. By making slight adjustments to the ego camera's angles in both positive and negative directions, CoDa-4DGS generates novel views that do not exist in the dataset. Additionally, it supports novel view synthesis for 4D semantic segmentation, offering a versatile tool for various scenarios. A detailed demonstration of this capability is included in the attached video.

%% file: sec/x5_discussion.tex
\section{Conceptual comparison}
\label{sec:discussion}

\setlength{\tabcolsep}{3pt} 
\begin{table*}[t!]
\centering
\fontsize{9}{12}\selectfont
\begin{threeparttable}
\caption{Conceptual comparison of very recent advancements in Gs-based scene rendering approaches.}
\label{table:dis_recentwork}
\begin{tabular}{L{3.2cm}C{2.1cm}C{2.1cm}C{2.1cm}C{2.1cm}C{2.1cm}}
\toprule
Approach &  4D Scene & Autonomous Driving  & Self-Supervised & Feature Distillation & Plug-and-Play \\ 
\midrule
Feature-3DGS~\papersource{(CVPR~24)}   &  & & \checkmark  &  \checkmark     \\
FMGS~\papersource{(IJCV~24)}   &  &  & \checkmark  &  \checkmark    \\
StreetGaussian~\papersource{(ECCV~24)}   & \checkmark  &\checkmark &   &    &   \\
$S^3$Gaussian~\papersource{(ECCV~24)}   & \checkmark &  \checkmark & \checkmark  &    &   \\
DN-4DGS~\papersource{(NeurIPS~24)}  &  \checkmark & & \checkmark &    &  \checkmark \\
\cellcolor{gray!20}\myMethod (Ours)  & \cellcolor{gray!20}\checkmark & \cellcolor{gray!20}\checkmark  & \cellcolor{gray!20}\checkmark   & \cellcolor{gray!20}\checkmark & \cellcolor{gray!20}\checkmark \\
\bottomrule
\end{tabular}
\end{threeparttable}
\end{table*}

To incorporate the latest advancements, we compare our method with $S^3$Gaussian~\cite{huang2024textit} and StreetGaussian~\cite{yan2024street}. Unlike StreetGaussian, which requires ground truth for training, our approach is built upon 4DGS and $S^3$Gaussian and leverages self-supervised learning. While ground truth offers precise tracking priors for dynamic objects in a scene, the self-supervised approach enhances scalability by eliminating the need for labeled bounding boxes. 

Additionally, it is worth emphasizing that our method can serve as a plug-and-play module to enhance other frameworks based on Gaussian temporal deformation. As demonstrated in Sec~\ref{sec:plugplay}, we focus on leveraging context awareness and deformation awareness to compensate for inaccuracies in vanilla Gaussian deformation predictions, thereby improving overall performance. For instance, a related method, DN-4DGS~\cite{lu2024dn}, also serves as a plug-and-play module and improves PSNR by 1.4\% for vanilla 4DGS. In comparison, our approach achieves an improvement of approximately 2\%. Moreover, DN-4DGS is not explicitly designed for autonomous driving scenes.

Using 2D foundation models to distill features has proven effective in 3DGS tasks~\cite{zhou2024feature, zuo2024fmgs}, especially with semantic information, which significantly aids scene understanding~\cite{kundu2022panoptic}. Semantic feature-supported 3DGS can enable downstream tasks like scene editing in static scenarios. We extend this idea to 4DGS, which is not straightforward and involves additional complexity. In 4DGS, each Gaussian undergoes temporal deformation, and the associated semantic features must also be transformed accordingly to maintain consistency in context after deformation. For example, if a Gaussian represents different objects at different time steps, its semantic features should adapt to reflect the new semantics, ensuring alignment between the rendered results and the foundation model's inferences. This consistency is essential for enabling \myMethod to perform semantic-aware applications, such as scene editor.

Furthermore, by incorporating semantic features as context-awareness inputs into the Deformable Compensation Network (DCN), we can constrain Gaussian deformation in spatial dimensions, thereby improving training outcomes. For instance, a Gaussian representing a road surface at one time step should remain consistent as a road surface after temporal deformation rather than erroneously transforming into a car due to proximity in 3D space. This is because the semantic feature distance between the two would be significantly larger despite their spatial proximity.

In Tab.~\ref{table:dis_recentwork}, we summarize the above discussions, highlighting the key distinctions between our approach and other very recent methods.

%% file: sec/x6_further_4Dvis.tex
\section{Further visual results}
\label{sec:further_vis}

\begin{figure*}[t!]
   \centering
   \includegraphics[width=0.95\textwidth]{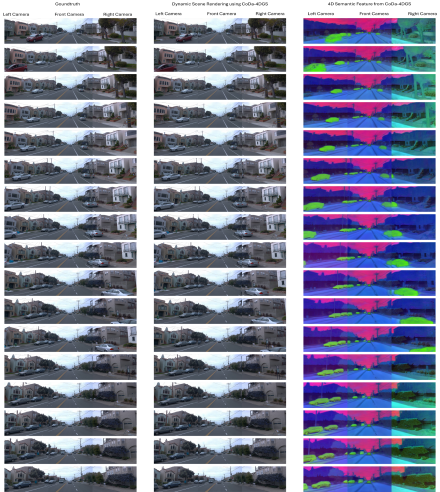}
   \caption{4D scene rendering for scene 86.}
   \label{fig:scene86}
\end{figure*}

\begin{figure*}[t!]
   \centering
   \includegraphics[width=0.95\textwidth]{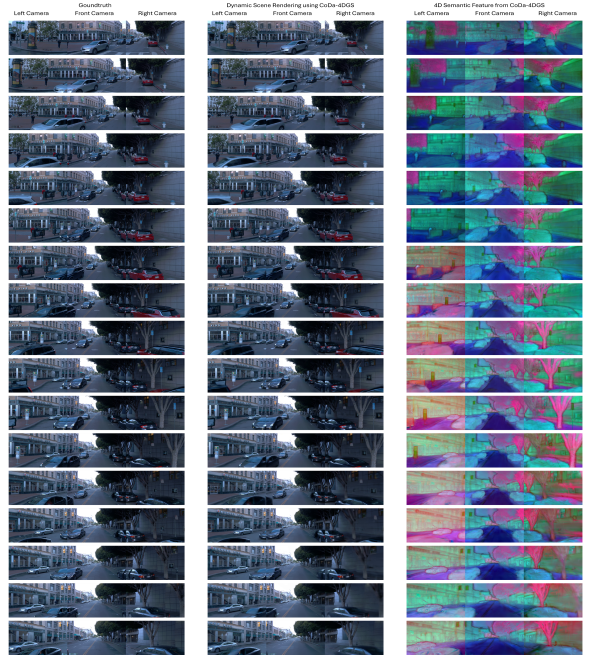}
   \caption{4D scene rendering for scene 80.}
   \label{fig:scene80}
\end{figure*}

\begin{figure*}[t!]
   \centering
   \includegraphics[width=0.95\textwidth]{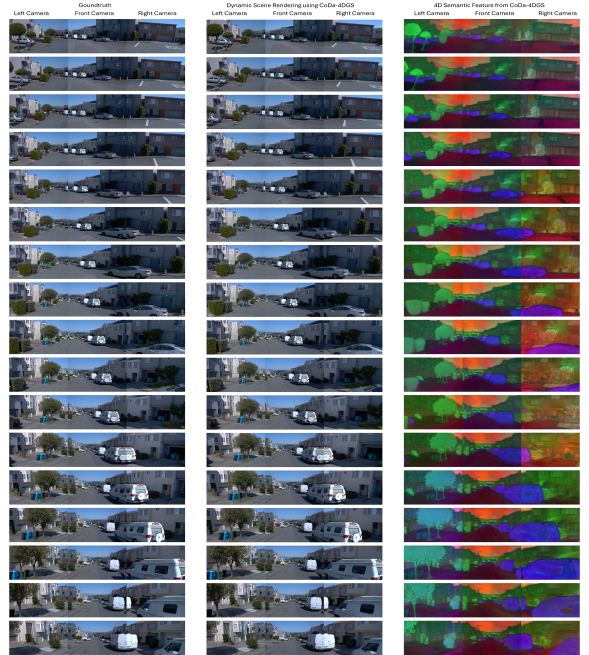}
   \caption{4D scene rendering for scene 03.}
   \label{fig:scene03}
\end{figure*}

\begin{figure*}[t!]
   \centering
   \includegraphics[width=0.95\textwidth]{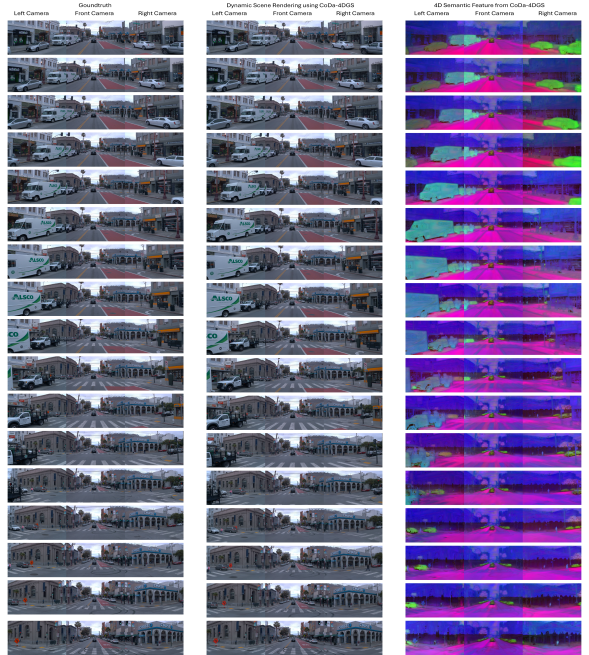}
   \caption{4D scene rendering for scene 22.}
   \label{fig:scene22}
\end{figure*}

To provide a more intuitive demonstration of CoDa-4DGS's performance on dynamic scenes, we present the RGB rendering results, semantic feature rendering results, and ground truth for each frame in chronological order. CoDa-4DGS consistently achieves exceptional 4D scene rendering across diverse scenarios, as shown in Scene 86 (Fig.\ref{fig:scene86}), Scene 80 (Fig.\ref{fig:scene80}), Scene 03 (Fig.\ref{fig:scene03}), and Scene 22 (Fig.\ref{fig:scene22}).

%% file: main.bbl
\begin{thebibliography}{67}
\providecommand{\natexlab}[1]{#1}
\providecommand{\url}[1]{\texttt{#1}}
\expandafter\ifx\csname urlstyle\endcsname\relax
  \providecommand{\doi}[1]{doi: #1}\else
  \providecommand{\doi}{doi: \begingroup \urlstyle{rm}\Url}\fi

\bibitem[Attal et~al.(2023)Attal, Huang, Richardt, Zollhoefer, Kopf, O’Toole, and Kim]{attal2023hyperreel}
Benjamin Attal, Jia-Bin Huang, Christian Richardt, Michael Zollhoefer, Johannes Kopf, Matthew O’Toole, and Changil Kim.
\newblock Hyperreel: High-fidelity 6-dof video with ray-conditioned sampling.
\newblock In \emph{Proceedings of the IEEE/CVF Conference on Computer Vision and Pattern Recognition}, pages 16610--16620, 2023.

\bibitem[Caesar et~al.(2021)Caesar, Kabzan, Tan, Fong, Wolff, Lang, Fletcher, Beijbom, and Omari]{caesar2021nuplan}
Holger Caesar, Juraj Kabzan, Kok~Seang Tan, Whye~Kit Fong, Eric Wolff, Alex Lang, Luke Fletcher, Oscar Beijbom, and Sammy Omari.
\newblock nuplan: A closed-loop ml-based planning benchmark for autonomous vehicles.
\newblock \emph{Computer Vision and Pattern Recognition (CVPR) 2021 ADP3 workshop}, 2021.

\bibitem[Cao and Johnson(2023)]{cao2023hexplane}
Ang Cao and Justin Johnson.
\newblock Hexplane: A fast representation for dynamic scenes.
\newblock In \emph{Proceedings of the IEEE/CVF Conference on Computer Vision and Pattern Recognition}, pages 130--141, 2023.

\bibitem[Chen et~al.(2022)Chen, Xu, Geiger, Yu, and Su]{Chen2022ECCV}
Anpei Chen, Zexiang Xu, Andreas Geiger, Jingyi Yu, and Hao Su.
\newblock Tensorf: Tensorial radiance fields.
\newblock In \emph{European Conference on Computer Vision (ECCV)}, 2022.

\bibitem[Dauner et~al.(2024)Dauner, Hallgarten, Li, Weng, Huang, Yang, Li, Gilitschenski, Ivanovic, Pavone, Geiger, and Chitta]{Dauner2024NEURIPS}
Daniel Dauner, Marcel Hallgarten, Tianyu Li, Xinshuo Weng, Zhiyu Huang, Zetong Yang, Hongyang Li, Igor Gilitschenski, Boris Ivanovic, Marco Pavone, Andreas Geiger, and Kashyap Chitta.
\newblock Navsim: Data-driven non-reactive autonomous vehicle simulation and benchmarking.
\newblock In \emph{Advances in Neural Information Processing Systems (NeurIPS)}, 2024.

\bibitem[Dosovitskiy and Brox(2016)]{dosovitskiy2016generating}
Alexey Dosovitskiy and Thomas Brox.
\newblock Generating images with perceptual similarity metrics based on deep networks.
\newblock \emph{Advances in neural information processing systems}, 29, 2016.

\bibitem[Dosovitskiy et~al.(2017)Dosovitskiy, Ros, Codevilla, Lopez, and Koltun]{dosovitskiy2017carla}
Alexey Dosovitskiy, German Ros, Felipe Codevilla, Antonio Lopez, and Vladlen Koltun.
\newblock Carla: An open urban driving simulator.
\newblock In \emph{Conference on robot learning}, pages 1--16. PMLR, 2017.

\bibitem[Fang et~al.(2022)Fang, Yi, Wang, Xie, Zhang, Liu, Nie{\ss}ner, and Tian]{fang2022fast}
Jiemin Fang, Taoran Yi, Xinggang Wang, Lingxi Xie, Xiaopeng Zhang, Wenyu Liu, Matthias Nie{\ss}ner, and Qi Tian.
\newblock Fast dynamic radiance fields with time-aware neural voxels.
\newblock In \emph{SIGGRAPH Asia 2022 Conference Papers}, pages 1--9, 2022.

\bibitem[Fischer et~al.(2024)Fischer, Kulhanek, Rota~Bul\`{o}, Porzi, Pollefeys, and Kontschieder]{NEURIPS2024_93397b48}
Tobias Fischer, Jonas Kulhanek, Samuel Rota~Bul\`{o}, Lorenzo Porzi, Marc Pollefeys, and Peter Kontschieder.
\newblock Dynamic 3d gaussian fields for urban areas.
\newblock In \emph{Advances in Neural Information Processing Systems}, pages 80466--80494. Curran Associates, Inc., 2024.

\bibitem[Fridovich-Keil et~al.(2023)Fridovich-Keil, Meanti, Warburg, Recht, and Kanazawa]{fridovich2023k}
Sara Fridovich-Keil, Giacomo Meanti, Frederik~Rahb{\ae}k Warburg, Benjamin Recht, and Angjoo Kanazawa.
\newblock K-planes: Explicit radiance fields in space, time, and appearance.
\newblock In \emph{Proceedings of the IEEE/CVF Conference on Computer Vision and Pattern Recognition}, pages 12479--12488, 2023.

\bibitem[Geiger et~al.(2013)Geiger, Lenz, Stiller, and Urtasun]{geiger2013vision}
Andreas Geiger, Philip Lenz, Christoph Stiller, and Raquel Urtasun.
\newblock Vision meets robotics: The kitti dataset.
\newblock \emph{The international journal of robotics research}, 32\penalty0 (11):\penalty0 1231--1237, 2013.

\bibitem[Guo et~al.(2023{\natexlab{a}})Guo, Deng, Li, Bai, Shi, Wang, Ding, Wang, and Li]{guo2023streetsurf}
Jianfei Guo, Nianchen Deng, Xinyang Li, Yeqi Bai, Botian Shi, Chiyu Wang, Chenjing Ding, Dongliang Wang, and Yikang Li.
\newblock Streetsurf: Extending multi-view implicit surface reconstruction to street views.
\newblock \emph{arXiv preprint arXiv:2306.04988}, 2023{\natexlab{a}}.

\bibitem[Guo et~al.(2023{\natexlab{b}})Guo, Sun, Dai, Chen, Ye, Tan, Ding, Zhang, and Wang]{guo2023forward}
Xiang Guo, Jiadai Sun, Yuchao Dai, Guanying Chen, Xiaoqing Ye, Xiao Tan, Errui Ding, Yumeng Zhang, and Jingdong Wang.
\newblock Forward flow for novel view synthesis of dynamic scenes.
\newblock In \emph{Proceedings of the IEEE/CVF International Conference on Computer Vision}, pages 16022--16033, 2023{\natexlab{b}}.

\bibitem[Huang et~al.(2024{\natexlab{a}})Huang, Wei, Zheng, An, Lu, Zhan, Tomizuka, Keutzer, and Zhang]{huang2024textit}
Nan Huang, Xiaobao Wei, Wenzhao Zheng, Pengju An, Ming Lu, Wei Zhan, Masayoshi Tomizuka, Kurt Keutzer, and Shanghang Zhang.
\newblock S3gaussian: Self-supervised street gaussians for autonomous driving.
\newblock In \emph{ECCV}, 2024{\natexlab{a}}.

\bibitem[Huang et~al.(2024{\natexlab{b}})Huang, Sun, Yang, Lyu, Cao, and Qi]{huang2024sc}
Yi-Hua Huang, Yang-Tian Sun, Ziyi Yang, Xiaoyang Lyu, Yan-Pei Cao, and Xiaojuan Qi.
\newblock Sc-gs: Sparse-controlled gaussian splatting for editable dynamic scenes.
\newblock In \emph{Proceedings of the IEEE/CVF Conference on Computer Vision and Pattern Recognition}, pages 4220--4230, 2024{\natexlab{b}}.

\bibitem[Hwang et~al.(2024)Hwang, Xu, Lin, Hung, Ji, Choi, Huang, He, Covington, Sapp, Zhou, Guo, Anguelov, and Tan]{hwang2024emma}
Jyh-Jing Hwang, Runsheng Xu, Hubert Lin, Wei-Chih Hung, Jingwei Ji, Kristy Choi, Di Huang, Tong He, Paul Covington, Benjamin Sapp, Yin Zhou, James Guo, Dragomir Anguelov, and Mingxing Tan.
\newblock Emma: End-to-end multimodal model for autonomous driving.
\newblock \emph{arXiv preprint arXiv:2410.23262}, 2024.

\bibitem[Jiang et~al.(2025)Jiang, Gao, Shao, Wang, Xiong, and Zhang]{10816486}
Changjian Jiang, Ruilan Gao, Kele Shao, Yue Wang, Rong Xiong, and Yu Zhang.
\newblock Li-gs: Gaussian splatting with lidar incorporated for accurate large-scale reconstruction.
\newblock \emph{IEEE Robotics and Automation Letters}, 10\penalty0 (2):\penalty0 1864--1871, 2025.

\bibitem[Kerbl et~al.(2023)Kerbl, Kopanas, Leimk{\"u}hler, and Drettakis]{kerbl3Dgaussians}
Bernhard Kerbl, Georgios Kopanas, Thomas Leimk{\"u}hler, and George Drettakis.
\newblock 3d gaussian splatting for real-time radiance field rendering.
\newblock \emph{ACM Transactions on Graphics}, 42\penalty0 (4), 2023.

\bibitem[Kirillov et~al.(2023)Kirillov, Mintun, Ravi, Mao, Rolland, Gustafson, Xiao, Whitehead, Berg, Lo, Doll{\'a}r, and Girshick]{kirillov2023segany}
Alexander Kirillov, Eric Mintun, Nikhila Ravi, Hanzi Mao, Chloe Rolland, Laura Gustafson, Tete Xiao, Spencer Whitehead, Alexander~C. Berg, Wan-Yen Lo, Piotr Doll{\'a}r, and Ross Girshick.
\newblock Segment anything.
\newblock \emph{arXiv:2304.02643}, 2023.

\bibitem[Kundu et~al.(2022)Kundu, Genova, Yin, Fathi, Pantofaru, Guibas, Tagliasacchi, Dellaert, and Funkhouser]{kundu2022panoptic}
Abhijit Kundu, Kyle Genova, Xiaoqi Yin, Alireza Fathi, Caroline Pantofaru, Leonidas~J Guibas, Andrea Tagliasacchi, Frank Dellaert, and Thomas Funkhouser.
\newblock Panoptic neural fields: A semantic object-aware neural scene representation.
\newblock In \emph{Proceedings of the IEEE/CVF Conference on Computer Vision and Pattern Recognition}, pages 12871--12881, 2022.

\bibitem[Li et~al.(2022{\natexlab{a}})Li, Weinberger, Belongie, Koltun, and Ranftl]{li2022language}
Boyi Li, Kilian~Q Weinberger, Serge Belongie, Vladlen Koltun, and Ren{\'e} Ranftl.
\newblock Language-driven semantic segmentation.
\newblock \emph{ICLR}, 2022{\natexlab{a}}.

\bibitem[Li et~al.(2022{\natexlab{b}})Li, Slavcheva, Zollhoefer, Green, Lassner, Kim, Schmidt, Lovegrove, Goesele, Newcombe, et~al.]{li2022neural}
Tianye Li, Mira Slavcheva, Michael Zollhoefer, Simon Green, Christoph Lassner, Changil Kim, Tanner Schmidt, Steven Lovegrove, Michael Goesele, Richard Newcombe, et~al.
\newblock Neural 3d video synthesis from multi-view video.
\newblock In \emph{Proceedings of the IEEE/CVF Conference on Computer Vision and Pattern Recognition}, pages 5521--5531, 2022{\natexlab{b}}.

\bibitem[Li et~al.(2021)Li, Niklaus, Snavely, and Wang]{li2021neural}
Zhengqi Li, Simon Niklaus, Noah Snavely, and Oliver Wang.
\newblock Neural scene flow fields for space-time view synthesis of dynamic scenes.
\newblock In \emph{Proceedings of the IEEE/CVF Conference on Computer Vision and Pattern Recognition}, pages 6498--6508, 2021.

\bibitem[Li et~al.(2023)Li, Wang, Cole, Tucker, and Snavely]{li2023dynibar}
Zhengqi Li, Qianqian Wang, Forrester Cole, Richard Tucker, and Noah Snavely.
\newblock Dynibar: Neural dynamic image-based rendering.
\newblock In \emph{Proceedings of the IEEE/CVF Conference on Computer Vision and Pattern Recognition}, pages 4273--4284, 2023.

\bibitem[Li et~al.(2024)Li, Chen, Li, and Xu]{Li_STG_2024_CVPR}
Zhan Li, Zhang Chen, Zhong Li, and Yi Xu.
\newblock Spacetime gaussian feature splatting for real-time dynamic view synthesis.
\newblock In \emph{Proceedings of the IEEE/CVF Conference on Computer Vision and Pattern Recognition (CVPR)}, pages 8508--8520, 2024.

\bibitem[Lin et~al.(2022)Lin, Peng, Xu, Yan, Shuai, Bao, and Zhou]{lin2022efficient}
Haotong Lin, Sida Peng, Zhen Xu, Yunzhi Yan, Qing Shuai, Hujun Bao, and Xiaowei Zhou.
\newblock Efficient neural radiance fields for interactive free-viewpoint video.
\newblock In \emph{SIGGRAPH Asia 2022 Conference Papers}, pages 1--9, 2022.

\bibitem[Lin et~al.(2023)Lin, Peng, Xu, Xie, He, Bao, and Zhou]{lin2023im4d}
Haotong Lin, Sida Peng, Zhen Xu, Tao Xie, Xingyi He, Hujun Bao, and Xiaowei Zhou.
\newblock High-fidelity and real-time novel view synthesis for dynamic scenes.
\newblock In \emph{SIGGRAPH Asia Conference Proceedings}, 2023.

\bibitem[Liu et~al.(2022)Liu, Cao, Mao, Zhang, Zhang, Keppo, Shan, Qie, and Shou]{liu2022devrf}
Jia-Wei Liu, Yan-Pei Cao, Weijia Mao, Wenqiao Zhang, David~Junhao Zhang, Jussi Keppo, Ying Shan, Xiaohu Qie, and Mike~Zheng Shou.
\newblock Devrf: Fast deformable voxel radiance fields for dynamic scenes.
\newblock \emph{arXiv preprint arXiv:2205.15723}, 2022.

\bibitem[Liu et~al.(2023)Liu, Chen, Yang, Wang, Manivasagam, and Urtasun]{liu2023real}
Jeffrey~Yunfan Liu, Yun Chen, Ze Yang, Jingkang Wang, Sivabalan Manivasagam, and Raquel Urtasun.
\newblock Real-time neural rasterization for large scenes.
\newblock In \emph{Proceedings of the IEEE/CVF International Conference on Computer Vision}, pages 8416--8427, 2023.

\bibitem[Ljungbergh et~al.(2024)Ljungbergh, Tonderski, Johnander, Caesar, {\AA}str{\"o}m, Felsberg, and Petersson]{ljungbergh2024neuroncap}
William Ljungbergh, Adam Tonderski, Joakim Johnander, Holger Caesar, Kalle {\AA}str{\"o}m, Michael Felsberg, and Christoffer Petersson.
\newblock Neuroncap: Photorealistic closed-loop safety testing for autonomous driving.
\newblock \emph{European Conference on Computer Vision (ECCV)}, 2024.

\bibitem[Lombardi et~al.(2021)Lombardi, Simon, Schwartz, Zollhoefer, Sheikh, and Saragih]{lombardi2021mixture}
Stephen Lombardi, Tomas Simon, Gabriel Schwartz, Michael Zollhoefer, Yaser Sheikh, and Jason Saragih.
\newblock Mixture of volumetric primitives for efficient neural rendering.
\newblock \emph{ACM Transactions on Graphics (ToG)}, 40\penalty0 (4):\penalty0 1--13, 2021.

\bibitem[Lu et~al.(2023)Lu, Xu, Chen, Li, Lin, and Jiang]{lu2023urban}
Fan Lu, Yan Xu, Guang Chen, Hongsheng Li, Kwan-Yee Lin, and Changjun Jiang.
\newblock Urban radiance field representation with deformable neural mesh primitives.
\newblock In \emph{Proceedings of the IEEE/CVF International Conference on Computer Vision}, pages 465--476, 2023.

\bibitem[Lu et~al.(2024)Lu, Deng, Zhu, Liang, Yang, Zhang, and Zhou]{lu2024dn}
Jiahao Lu, Jiacheng Deng, Ruijie Zhu, Yanzhe Liang, Wenfei Yang, Tianzhu Zhang, and Xu Zhou.
\newblock Dn-4dgs: Denoised deformable network with temporal-spatial aggregation for dynamic scene rendering.
\newblock In \emph{NeurIPS}, 2024.

\bibitem[Luiten et~al.(2024)Luiten, Kopanas, Leibe, and Ramanan]{luiten2024dynamic}
Jonathon Luiten, Georgios Kopanas, Bastian Leibe, and Deva Ramanan.
\newblock Dynamic 3d gaussians: Tracking by persistent dynamic view synthesis.
\newblock In \emph{2024 International Conference on 3D Vision (3DV)}, pages 800--809. IEEE, 2024.

\bibitem[Ost et~al.(2021)Ost, Mannan, Thuerey, Knodt, and Heide]{ost2021neural}
Julian Ost, Fahim Mannan, Nils Thuerey, Julian Knodt, and Felix Heide.
\newblock Neural scene graphs for dynamic scenes.
\newblock In \emph{Proceedings of the IEEE/CVF Conference on Computer Vision and Pattern Recognition}, pages 2856--2865, 2021.

\bibitem[Park et~al.(2021{\natexlab{a}})Park, Sinha, Barron, Bouaziz, Goldman, Seitz, and Martin-Brualla]{park2021nerfies}
Keunhong Park, Utkarsh Sinha, Jonathan~T Barron, Sofien Bouaziz, Dan~B Goldman, Steven~M Seitz, and Ricardo Martin-Brualla.
\newblock Nerfies: Deformable neural radiance fields.
\newblock In \emph{Proceedings of the IEEE/CVF International Conference on Computer Vision}, pages 5865--5874, 2021{\natexlab{a}}.

\bibitem[Park et~al.(2021{\natexlab{b}})Park, Sinha, Hedman, Barron, Bouaziz, Goldman, Martin-Brualla, and Seitz]{park2021hypernerf}
Keunhong Park, Utkarsh Sinha, Peter Hedman, Jonathan~T. Barron, Sofien Bouaziz, Dan~B Goldman, Ricardo Martin-Brualla, and Steven~M. Seitz.
\newblock Hypernerf: A higher-dimensional representation for topologically varying neural radiance fields.
\newblock \emph{ACM Trans. Graph.}, 40\penalty0 (6), 2021{\natexlab{b}}.

\bibitem[Peng et~al.(2023)Peng, Yan, Shuai, Bao, and Zhou]{peng2023representing}
Sida Peng, Yunzhi Yan, Qing Shuai, Hujun Bao, and Xiaowei Zhou.
\newblock Representing volumetric videos as dynamic mlp maps.
\newblock In \emph{Proceedings of the IEEE/CVF Conference on Computer Vision and Pattern Recognition}, pages 4252--4262, 2023.

\bibitem[Pumarola et~al.(2021)Pumarola, Corona, Pons-Moll, and Moreno-Noguer]{pumarola2021d}
Albert Pumarola, Enric Corona, Gerard Pons-Moll, and Francesc Moreno-Noguer.
\newblock D-nerf: Neural radiance fields for dynamic scenes.
\newblock In \emph{Proceedings of the IEEE/CVF Conference on Computer Vision and Pattern Recognition}, pages 10318--10327, 2021.

\bibitem[Rematas et~al.(2022)Rematas, Liu, Srinivasan, Barron, Tagliasacchi, Funkhouser, and Ferrari]{rematas2022urban}
Konstantinos Rematas, Andrew Liu, Pratul~P Srinivasan, Jonathan~T Barron, Andrea Tagliasacchi, Thomas Funkhouser, and Vittorio Ferrari.
\newblock Urban radiance fields.
\newblock In \emph{Proceedings of the IEEE/CVF Conference on Computer Vision and Pattern Recognition}, pages 12932--12942, 2022.

\bibitem[Rudnev et~al.(2022)Rudnev, Elgharib, Smith, Liu, Golyanik, and Theobalt]{rudnev2022nerf}
Viktor Rudnev, Mohamed Elgharib, William Smith, Lingjie Liu, Vladislav Golyanik, and Christian Theobalt.
\newblock Nerf for outdoor scene relighting.
\newblock In \emph{European Conference on Computer Vision}, pages 615--631. Springer, 2022.

\bibitem[Schonberger and Frahm(2016)]{schonberger2016structure}
Johannes~L Schonberger and Jan-Michael Frahm.
\newblock Structure-from-motion revisited.
\newblock In \emph{Proceedings of the IEEE conference on computer vision and pattern recognition}, pages 4104--4113, 2016.

\bibitem[Song et~al.(2023)Song, Chen, Li, Chen, Chen, Yuan, Xu, and Geiger]{song2023nerfplayer}
Liangchen Song, Anpei Chen, Zhong Li, Zhang Chen, Lele Chen, Junsong Yuan, Yi Xu, and Andreas Geiger.
\newblock Nerfplayer: A streamable dynamic scene representation with decomposed neural radiance fields.
\newblock \emph{IEEE Transactions on Visualization and Computer Graphics}, 29\penalty0 (5):\penalty0 2732--2742, 2023.

\bibitem[Sun et~al.(2020)Sun, Kretzschmar, Dotiwalla, Chouard, Patnaik, Tsui, Guo, Zhou, Chai, Caine, et~al.]{sun2020scalability}
Pei Sun, Henrik Kretzschmar, Xerxes Dotiwalla, Aurelien Chouard, Vijaysai Patnaik, Paul Tsui, James Guo, Yin Zhou, Yuning Chai, Benjamin Caine, et~al.
\newblock Scalability in perception for autonomous driving: Waymo open dataset.
\newblock In \emph{Proceedings of the IEEE/CVF conference on computer vision and pattern recognition}, pages 2446--2454, 2020.

\bibitem[Sun et~al.(2024)Sun, Zhuang, Jiang, Liu, Xie, and Chandraker]{sun2024lidarf}
Shanlin Sun, Bingbing Zhuang, Ziyu Jiang, Buyu Liu, Xiaohui Xie, and Manmohan Chandraker.
\newblock Lidarf: Delving into lidar for neural radiance field on street scenes.
\newblock In \emph{Proceedings of the IEEE/CVF Conference on Computer Vision and Pattern Recognition}, pages 19563--19572, 2024.

\bibitem[Tancik et~al.(2022)Tancik, Casser, Yan, Pradhan, Mildenhall, Srinivasan, Barron, and Kretzschmar]{tancik2022block}
Matthew Tancik, Vincent Casser, Xinchen Yan, Sabeek Pradhan, Ben Mildenhall, Pratul~P Srinivasan, Jonathan~T Barron, and Henrik Kretzschmar.
\newblock Block-nerf: Scalable large scene neural view synthesis.
\newblock In \emph{Proceedings of the IEEE/CVF Conference on Computer Vision and Pattern Recognition}, pages 8248--8258, 2022.

\bibitem[Tonderski et~al.(2024)Tonderski, Lindstr{\"o}m, Hess, Ljungbergh, Svensson, and Petersson]{tonderski2024neurad}
Adam Tonderski, Carl Lindstr{\"o}m, Georg Hess, William Ljungbergh, Lennart Svensson, and Christoffer Petersson.
\newblock Neurad: Neural rendering for autonomous driving.
\newblock In \emph{Proceedings of the IEEE/CVF Conference on Computer Vision and Pattern Recognition}, pages 14895--14904, 2024.

\bibitem[Tretschk et~al.(2021)Tretschk, Tewari, Golyanik, Zollh{\"o}fer, Lassner, and Theobalt]{tretschk2021non}
Edgar Tretschk, Ayush Tewari, Vladislav Golyanik, Michael Zollh{\"o}fer, Christoph Lassner, and Christian Theobalt.
\newblock Non-rigid neural radiance fields: Reconstruction and novel view synthesis of a dynamic scene from monocular video.
\newblock In \emph{Proceedings of the IEEE/CVF International Conference on Computer Vision}, pages 12959--12970, 2021.

\bibitem[Turki et~al.(2022)Turki, Ramanan, and Satyanarayanan]{turki2022mega}
Haithem Turki, Deva Ramanan, and Mahadev Satyanarayanan.
\newblock Mega-nerf: Scalable construction of large-scale nerfs for virtual fly-throughs.
\newblock In \emph{Proceedings of the IEEE/CVF Conference on Computer Vision and Pattern Recognition}, pages 12922--12931, 2022.

\bibitem[Turki et~al.(2023)Turki, Zhang, Ferroni, and Ramanan]{turki2023suds}
Haithem Turki, Jason~Y Zhang, Francesco Ferroni, and Deva Ramanan.
\newblock Suds: Scalable urban dynamic scenes.
\newblock In \emph{Proceedings of the IEEE/CVF Conference on Computer Vision and Pattern Recognition}, pages 12375--12385, 2023.

\bibitem[Wang et~al.(2023{\natexlab{a}})Wang, Tan, Li, Tian, Song, and Liu]{wang2023mixed}
Feng Wang, Sinan Tan, Xinghang Li, Zeyue Tian, Yafei Song, and Huaping Liu.
\newblock Mixed neural voxels for fast multi-view video synthesis.
\newblock In \emph{Proceedings of the IEEE/CVF International Conference on Computer Vision}, pages 19706--19716, 2023{\natexlab{a}}.

\bibitem[Wang et~al.(2023{\natexlab{b}})Wang, Hu, He, Wang, Yu, Tuytelaars, Xu, and Wu]{wang2023neural}
Liao Wang, Qiang Hu, Qihan He, Ziyu Wang, Jingyi Yu, Tinne Tuytelaars, Lan Xu, and Minye Wu.
\newblock Neural residual radiance fields for streamably free-viewpoint videos.
\newblock In \emph{Proceedings of the IEEE/CVF Conference on Computer Vision and Pattern Recognition}, pages 76--87, 2023{\natexlab{b}}.

\bibitem[Wang et~al.(2004)Wang, Bovik, Sheikh, and Simoncelli]{wang2004image}
Zhou Wang, Alan~C Bovik, Hamid~R Sheikh, and Eero~P Simoncelli.
\newblock Image quality assessment: from error visibility to structural similarity.
\newblock \emph{IEEE transactions on image processing}, 13\penalty0 (4):\penalty0 600--612, 2004.

\bibitem[Wu et~al.(2024)Wu, Yi, Fang, Xie, Zhang, Wei, Liu, Tian, and Wang]{wu20244d}
Guanjun Wu, Taoran Yi, Jiemin Fang, Lingxi Xie, Xiaopeng Zhang, Wei Wei, Wenyu Liu, Qi Tian, and Xinggang Wang.
\newblock 4d gaussian splatting for real-time dynamic scene rendering.
\newblock In \emph{Proceedings of the IEEE/CVF Conference on Computer Vision and Pattern Recognition}, pages 20310--20320, 2024.

\bibitem[Wu et~al.(2023)Wu, Liu, Luo, Zhong, Chen, Xiao, Hou, Lou, Chen, Yang, et~al.]{wu2023mars}
Zirui Wu, Tianyu Liu, Liyi Luo, Zhide Zhong, Jianteng Chen, Hongmin Xiao, Chao Hou, Haozhe Lou, Yuantao Chen, Runyi Yang, et~al.
\newblock Mars: An instance-aware, modular and realistic simulator for autonomous driving.
\newblock In \emph{CAAI International Conference on Artificial Intelligence}, pages 3--15. Springer, 2023.

\bibitem[Xian et~al.(2021)Xian, Huang, Kopf, and Kim]{xian2021space}
Wenqi Xian, Jia-Bin Huang, Johannes Kopf, and Changil Kim.
\newblock Space-time neural irradiance fields for free-viewpoint video.
\newblock In \emph{Proceedings of the IEEE/CVF conference on computer vision and pattern recognition}, pages 9421--9431, 2021.

\bibitem[Xie et~al.(2023)Xie, Zhang, Li, Zhang, and Zhang]{xie2023snerf}
Ziyang Xie, Junge Zhang, Wenye Li, Feihu Zhang, and Li Zhang.
\newblock S-nerf: Neural radiance fields for street views.
\newblock In \emph{International Conference on Learning Representations (ICLR)}, 2023.

\bibitem[Yan et~al.(2024)Yan, Lin, Zhou, Wang, Sun, Zhan, Lang, Zhou, and Peng]{yan2024street}
Yunzhi Yan, Haotong Lin, Chenxu Zhou, Weijie Wang, Haiyang Sun, Kun Zhan, Xianpeng Lang, Xiaowei Zhou, and Sida Peng.
\newblock Street gaussians for modeling dynamic urban scenes.
\newblock In \emph{European Conference on Computer Vision}, 2024.

\bibitem[Yang et~al.(2023{\natexlab{a}})Yang, Ivanovic, Litany, Weng, Kim, Li, Che, Xu, Fidler, Pavone, and Wang]{yang2023emernerf}
Jiawei Yang, Boris Ivanovic, Or Litany, Xinshuo Weng, Seung~Wook Kim, Boyi Li, Tong Che, Danfei Xu, Sanja Fidler, Marco Pavone, and Yue Wang.
\newblock Emernerf: Emergent spatial-temporal scene decomposition via self-supervision.
\newblock \emph{arXiv preprint arXiv:2311.02077}, 2023{\natexlab{a}}.

\bibitem[Yang et~al.(2023{\natexlab{b}})Yang, Chen, Wang, Manivasagam, Ma, Yang, and Urtasun]{yang2023unisim}
Ze Yang, Yun Chen, Jingkang Wang, Sivabalan Manivasagam, Wei-Chiu Ma, Anqi~Joyce Yang, and Raquel Urtasun.
\newblock Unisim: A neural closed-loop sensor simulator.
\newblock In \emph{Proceedings of the IEEE/CVF Conference on Computer Vision and Pattern Recognition}, pages 1389--1399, 2023{\natexlab{b}}.

\bibitem[Yang et~al.(2024{\natexlab{a}})Yang, Gao, Zhou, Jiao, Zhang, and Jin]{yang2024deformable}
Ziyi Yang, Xinyu Gao, Wen Zhou, Shaohui Jiao, Yuqing Zhang, and Xiaogang Jin.
\newblock Deformable 3d gaussians for high-fidelity monocular dynamic scene reconstruction.
\newblock In \emph{Proceedings of the IEEE/CVF Conference on Computer Vision and Pattern Recognition}, pages 20331--20341, 2024{\natexlab{a}}.

\bibitem[Yang et~al.(2024{\natexlab{b}})Yang, Yang, Pan, and Zhang]{yang2024gs4d}
Zeyu Yang, Hongye Yang, Zijie Pan, and Li Zhang.
\newblock Real-time photorealistic dynamic scene representation and rendering with 4d gaussian splatting.
\newblock In \emph{International Conference on Learning Representations (ICLR)}, 2024{\natexlab{b}}.

\bibitem[Zhou et~al.(2024{\natexlab{a}})Zhou, Shao, Xu, Bai, Qiu, Liu, Wang, Geiger, and Liao]{zhou2024hugs}
Hongyu Zhou, Jiahao Shao, Lu Xu, Dongfeng Bai, Weichao Qiu, Bingbing Liu, Yue Wang, Andreas Geiger, and Yiyi Liao.
\newblock Hugs: Holistic urban 3d scene understanding via gaussian splatting.
\newblock In \emph{Proceedings of the IEEE/CVF Conference on Computer Vision and Pattern Recognition}, pages 21336--21345, 2024{\natexlab{a}}.

\bibitem[Zhou et~al.(2024{\natexlab{b}})Zhou, Chang, Jiang, Fan, Zhu, Xu, Chari, You, Wang, and Kadambi]{zhou2024feature}
Shijie Zhou, Haoran Chang, Sicheng Jiang, Zhiwen Fan, Zehao Zhu, Dejia Xu, Pradyumna Chari, Suya You, Zhangyang Wang, and Achuta Kadambi.
\newblock Feature 3dgs: Supercharging 3d gaussian splatting to enable distilled feature fields.
\newblock In \emph{Proceedings of the IEEE/CVF Conference on Computer Vision and Pattern Recognition}, pages 21676--21685, 2024{\natexlab{b}}.

\bibitem[Zhou et~al.(2024{\natexlab{c}})Zhou, Lin, Shan, Wang, Sun, and Yang]{zhou2024drivinggaussian}
Xiaoyu Zhou, Zhiwei Lin, Xiaojun Shan, Yongtao Wang, Deqing Sun, and Ming-Hsuan Yang.
\newblock Drivinggaussian: Composite gaussian splatting for surrounding dynamic autonomous driving scenes.
\newblock In \emph{Proceedings of the IEEE/CVF Conference on Computer Vision and Pattern Recognition}, pages 21634--21643, 2024{\natexlab{c}}.

\bibitem[Zhu et~al.(2024)Zhu, Liang, Chang, Deng, Lu, Yang, Zhang, and Zhang]{zhu2024motiongs}
Ruijie Zhu, Yanzhe Liang, Hanzhi Chang, Jiacheng Deng, Jiahao Lu, Wenfei Yang, Tianzhu Zhang, and Yongdong Zhang.
\newblock Motiongs: Exploring explicit motion guidance for deformable 3d gaussian splatting.
\newblock In \emph{NeurIPS}, 2024.

\bibitem[Zuo et~al.(2024)Zuo, Samangouei, Zhou, Di, and Li]{zuo2024fmgs}
Xingxing Zuo, Pouya Samangouei, Yunwen Zhou, Yan Di, and Mingyang Li.
\newblock Fmgs: Foundation model embedded 3d gaussian splatting for holistic 3d scene understanding.
\newblock \emph{International Journal of Computer Vision}, pages 1--17, 2024.

\end{thebibliography}
